\documentclass{article}

    \PassOptionsToPackage{numbers, compress}{natbib}

\usepackage[final]{neurips_2025}

\usepackage[utf8]{inputenc} 
\usepackage[T1]{fontenc}    
\usepackage{hyperref}       
\usepackage{url}            
\usepackage{booktabs}       
\usepackage{amsfonts,amsmath,amsthm}       
\usepackage{nicefrac}       
\usepackage{microtype}      
\usepackage{xcolor}         
\usepackage{algorithm,algorithmic}
\usepackage{graphicx}
\usepackage{subcaption}
\usepackage{wrapfig}
\usepackage{todonotes}

\def\SS{\mathcal{S}}
\def\A{\mathcal{A}}
\def\Z{\mathcal{Z}}
\def\L{\mathcal{L}}
\def\E{\mathbb{E}}

\def\H{\mathcal{H}}
\def\RNS{\mathbb{R}}

\def\gm{\mathrm{GM}}
\def\de{\mathrm{d}}
\def\mse{\mathrm{MSE}}
\def\gaus{\mathcal{N}}
\def\piold{\mathcal{\pi_\text{old}}}
\def\pinew{\mathcal{\pi_\text{new}}}
\def\argmax{\mathop{\mathrm{argmax}}}
\def\argmin{\mathop{\mathrm{argmin}}}

\newcommand{\piat}[1]{\pi(\cdot|#1)}
\newcommand{\pitat}[1]{\pi_\theta(\cdot|#1)}

\title{Learning Intractable Multimodal Policies with Reparameterization and Diversity Regularization}

\author{%
  Ziqi Wang $\quad\quad$ Jiashun Liu $\quad\quad$ Ling Pan\thanks{Correspondence to: Ling Pan (lingpan@ust.hk)} \\\\
  Hong Kong University of Science and Technology
}

\begin{document}

\maketitle

\begin{abstract}
    Traditional continuous deep reinforcement learning (RL) algorithms employ deterministic or unimodal Gaussian actors, which cannot express complex multimodal decision distributions. This limitation can hinder their performance in diversity-critical scenarios.
    There have been some attempts to design online multimodal RL algorithms based on diffusion or amortized actors. However, these actors are intractable, making existing methods struggle with balancing performance, decision diversity, and efficiency simultaneously.
    To overcome this challenge, we first reformulate existing intractable multimodal actors within a unified framework, and prove that they can be directly optimized by policy gradient via reparameterization. 
    Then, we propose a distance-based diversity regularization that does not explicitly require decision probabilities.
    We identify two diversity-critical domains, namely multi-goal achieving and generative RL, to demonstrate the advantages of multimodal policies and our method, particularly in terms of few-shot robustness. 
    In conventional MuJoCo benchmarks, our algorithm also shows competitive performance. Moreover, our experiments highlight that the amortized actor is a promising policy model class with strong multimodal expressivity and high performance. Our code is available at \url{https://github.com/PneuC/DrAC}  
\end{abstract}
\section{Introduction}
\begin{wrapfigure}{r}{0.3\linewidth}
    \vspace{-1.15em}
    \includegraphics[width=\linewidth]{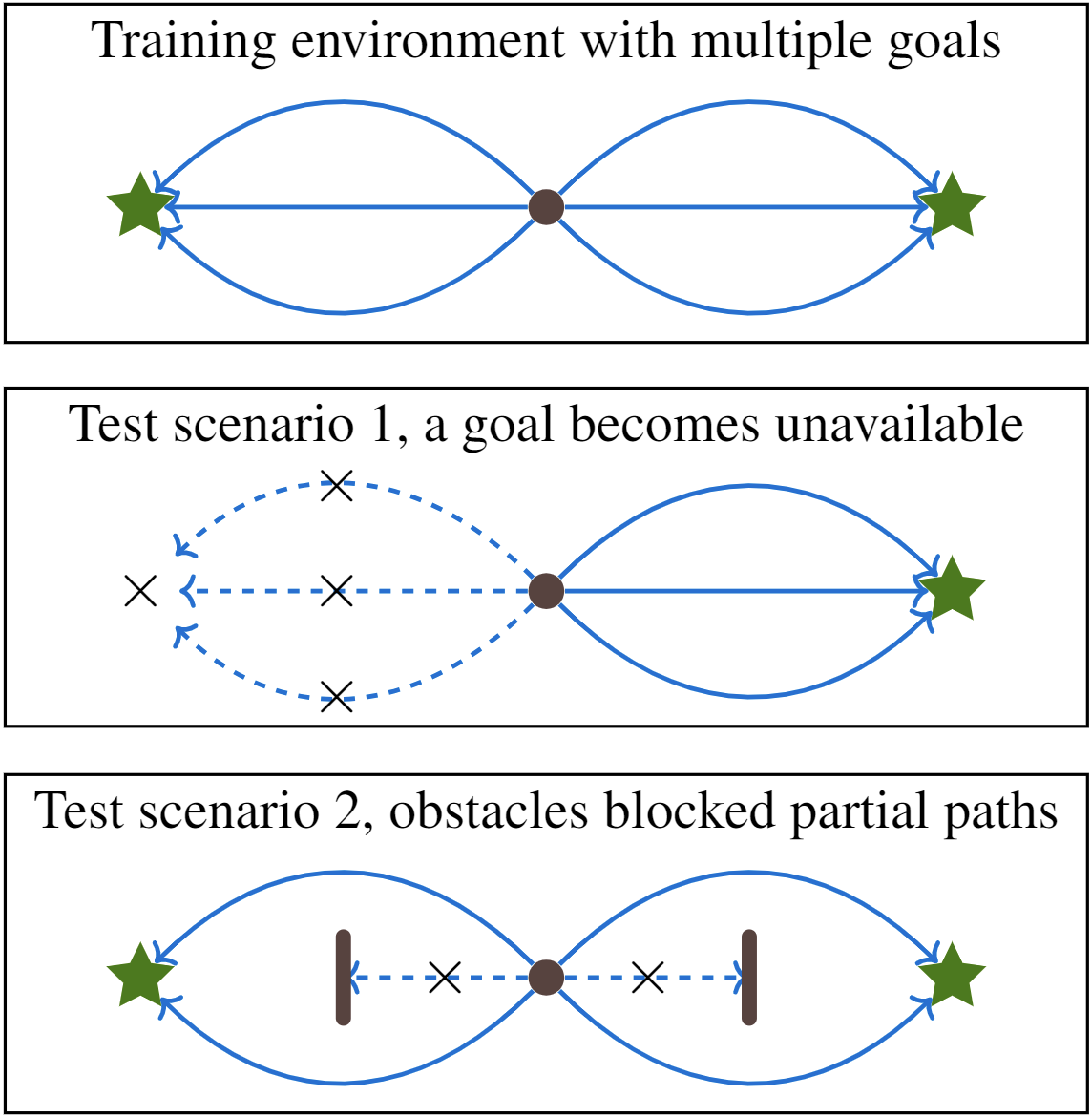}
    \caption{Motivational examples. A multimodal policy with maximized diversity can achieve multiple goals and enable robustness against environmental changes.}
    \vspace{-2em}
\end{wrapfigure}
Despite the remarkable progress in reinforcement learning (RL) for continuous control and decision-making tasks \cite{jumper2021highly,yarats2022mastering,tang2025deep}, learning multimodal policies remains challenging \cite{huang2023reparameterized,messaoud2024s,wang2024diffusion}. 
However, most state-of-the-art RL algorithms predominantly employ deterministic or unimodal Gaussian policies \cite{lillicrap2016continuous,schulman2017proximal,fujimoto2018addressing,haarnoja2018soft}, which limits their ability to capture the complex, multimodal decision distributions. 
However, real-world applications can have multiple goals, making multimodality and decision diversity essential \cite{haarnoja2017reinforcement,li2024learning,sutton1998reinforcement}, where unimodal policies can be brittle to perturbations in the environment. For example, given a navigation task in a maze with multiple valid paths, unimodal policies typically converge to a single shortest route, creating a critical vulnerability -- when this route becomes obstructed during deployment, such policies fail catastrophically~\cite{kumar2020one}. In contrast, a multimodal policy with high decision diversity can learn varied navigation strategies, allowing it to adapt when faced with unexpected obstacles. 
In zero-sum games, strategic diversity that requires multimodal policies is also critical for agents to maintain robustness against adaptive opponents \cite{liu2021towards}.
Additionally, RL has been applied to generative tasks in recent years \cite{cao2023reinforcement}, where multimodal policies can potentially improve the diversity of generated objects with less harm to quality \cite{wang2024negatively}.

The main challenge of learning expressive multimodal policies lies in their intractability. 
Existing online RL algorithms that train multimodal policies are typically based on the maximum-entropy RL framework \cite{haarnoja2018soft}. However, entropy regularization encourages policies to be more uniform, but do not specifically prefer multimodality. Meawhile, computing entropy requires an analytical expression of decision probability, which is not directly applicable to intractable actors. Some prior works attempt to mitigate this problem with a workaround that employs tractable but less expressive multimodal actors \cite{mazoure2020leveraging,cetin2022policy,wang2024negatively}, sacrificing expressivity. Another line of research leverages Stein variational gradient descent (SVGD) as a gradient estimator \cite{haarnoja2017reinforcement} or as a sampler that directly samples actions from the value distribution learned by the critic \cite{messaoud2024s}. However,  we empirically observe that it either underperforms traditional algorithms or incurs significant computation costs. 
Besides, there have been some online RL algorithms training diffusion actors \cite{yang2023policy,psenka2024learning,wang2024diffusion}, but to our best knowledge, only the diffusion actor-critic with entropy regulator (DACER) \cite{wang2024diffusion} takes decision diversity into account among these algorithms. DACER controls diversity by scaling an additional unimodal Gaussian noise instead of optimizing diversity through gradients w.r.t. the actor's parameters.

In this paper, we propose a novel diversity-regularized RL framework to address the aforementioned challenge. First, this paper shows that existing multimodal actors can be viewed as a combination of a parameterized mapping function with a fixed latent random distribution, which we call \textit{stochastic-mapping actors}. 
We then highlight that the policy gradient of any stochastic-mapping actor can be estimated via the reparameterization trick, without accessing the gradient of decision probabilities. This provides a general methodology to optimize intractable stochastic-mapping actors.
Secondly, we propose a distance-based diversity regularization instead of entropy, which does not explicitly require the decision probability. Combining with the policy regularization theorems \cite{wang2024diffusion}, we develop a novelty actor-critic algorithm named \textit{\underline{D}iversity-\underline{r}egularized \underline{A}ctor \underline{C}ritic} (DrAC).
Main contributions of this paper are summarized as follows
\begin{enumerate}
    \item We formulate a class of actors named \textit{stochastic-mapping actors}, bridge policy gradient and intractable stochastic actors via reparameterization trick. This formulation serves as a useful tool for understanding and designing multimodal RL models and algorithms.
    \item We propose a learning framework with a distance-based diversity regularization, which simultaneously optimizes expected return and decision diversity without requiring explicit access to decision probabilities.
    \item We identify two domains where multimodality and diversity are critical, and conduct comprehensive experiments, including performance in standard MuJoCo benchmarks. In addition to justifying the advantages of multimodal policies and our algorithm, we also present new insights on multimodality and diversity.
\end{enumerate}

\section{Related Works}
\paragraph{Policy Diversity in RL}

Decision diversity (or randomness) is an essential ingredient of online deep RL algorithms \cite{lillicrap2016continuous,fujimoto2018addressing,haarnoja2018soft}, which not only enforces exploration but also reduces variance \cite{fujimoto2018addressing}. Some research has shown that decision diversity can improve sample efficiency or final performance regarding expected return \cite{han2021diversity,parker2020effective,hong2018diversity,yang2022towards,conti2018improving}. 
Furthermore, diversity can serve in pretraining. Eysenbach et al. proposed that training policies with only a diversity objective can discover useful skills and serve as an effective pretraining method \cite{eysenbach2019diversity}. Ying et al. pretrain diffusion actors with intrinsic reward to discover diverse behaviors and fine-tune diffusion actors to fast generalize to downstream tasks  \cite{ying2025exploratory}.
Besides, Zhang et al. propose an algorithm to find varied policies for the same task \cite{zhang2019learning}. Kumar et al. train multiple policies with diverse behaviors, and highlight that diversity is the key to ensuring few-shot robustness in out-of-distribution scenarios \cite{kumar2020one}.

Another category of research on policy diversity in RL considers diversity as an independent objective distinguished from reward.
An emerging topic called quality-diversity RL aims at learning a set of policies that cover a task-specified behavior space as completely as possible, and maximizing the quality of each policy at the same time \cite{hegde2023generating,wu2023quality,xue2024sample}.
RL is also applied in generative tasks, including code generation, music generation, artificial intelligence for science, and game content generation \cite{cao2023reinforcement}. Some of these generative tasks treat diversity as an essential need besides quality. For example, Popova et al. formulate drug design as a Markov decision process (MDP) rewarded by a predictive model to generate diverse new chemical structures \cite{popova2018deep}.
Wang et al. leverage diversity-driven ensemble RL to generate diverse game levels with high quality \cite{wang2024negatively}.

\paragraph{Learning Multimodal Policies} An early attempt in online multimodal RL is soft Q-learning (SQL) \cite{haarnoja2017reinforcement}. It proposes the maximum-entropy RL framework with an amortized actor. As the actor loss in maximum-entropy RL is a KL-divergence, SQL leverages SVGD to estimate the gradient of the actor loss. Stein soft actor-critic (S$^2$AC) \cite{messaoud2024s} views the entropy-regularized critic as an energy that directly samples actions based on it via SVGD. To improve inference speed, S$^2$AC proposes to train an additional amortized actor that mimics the SVGD sampler. However, our experiments show that SQL performs poorly in many benchmarks, including MuJoCo, while training S$^2$AC is much more memory-hungry and slower. Huang et al. propose a model-based RL method learning through evidence lower bound to train multimodal actors \cite{huang2023reparameterized}. 
To our best knowledge, a fast and effective algorithm to train an amortized actor was absent prior to our work.

Another line of multimodal RL builds upon diffusion actors. There has been a lot of work investigating offline RL based on diffusion actor \cite{zhu2023diffusion,wang2023diffusion,kang2023efficient,cheng2024aligning,chen2024score}, but online methods have been a few. 
Specifically, Yang et al. \cite{yang2023policy} apply approximated policy improvement to generate target actions and train the diffusion actor to match the target actions. 
Psenka et al. \cite{psenka2024learning} train the diffusion actor by matching the score-based structure with the action gradient of the Q-function. Ren et al. model the diffusion process as a low-level MDP and train the diffusion actor through a two-stage policy gradient \cite{ren2024diffusion}.
Consistency policy methods have been demonstrated feasible to train diffusion actors online \cite{ding2024consistency,chen2024boosting}.
Wang et al. propose DACER, an online RL algorithm with entropy regulator to train diffusion actor \cite{wang2024diffusion}, taking decision diversity into consideration by automatically tuning the scale of an additional action noise. In addition, Ying et al. pre-train a Gaussian actor with intrinsic reward \cite{ying2025exploratory} to collect trajectories, and then fine-tune a diffusion actor in downstream tasks. Li et al. cluster the trajectories explored by learning with intrinsic rewards to discover modes, then utilize behavior cloning to train a diffusion actor \cite{li2024learning}. Jain et al. propose an energy-based training method for diffusion actors \cite{jain2025sampling}. Ishfaq et al. employ a diffusion model to generate synthetic training samples \cite{ishfaq2025langevin}.

There has also been some work on leveraging tractable multimodal policy models \cite{ren2021probabilistic,wang2024negatively}. Mazoure et al. leverage normalizing flow to improve exploration \cite{mazoure2020leveraging}. Wang et al. represent decision distribution by Gaussian mixture model to generate diverse game levels \cite{wang2024negatively}. However, these models may be weaker in expressivity compared with intractable models.

\section{Preliminaries}

\subsection{Deep Reinforcement Learning}
Deep reinforcement learning \cite{sutton1998reinforcement} aims at optimizing a policy $\pi$ powered by deep learning to solve an MDP. An MDP involves a state space $\SS$, an action space $\A$, a reward function $r(\cdot, \cdot): \SS \times \A \mapsto \RNS$, and a transition dynamics modeling the transition probability from one state to another given an action. Specifically, this paper focuses on the continuous action space.
At each time step $t$, the policy observes a state $S_t \in \SS$ to determine a decision distribution $\piat{S_t}$ and draw an action $A_t \in \A $ from $\piat{S_t}$, and then receives a reward signal $R_t = r(S_t, A_t)$.
The objective of RL is optimizing $\pi$ to maximize the expected return $J(\pi) = \E_{\pi}[\sum_{t=0}^\infty \gamma^t R_t].$  
A state-action value function conditioned by a policy $\pi$ is defined as $Q^\pi(s, a) = \E_\pi[\sum_{k=0}^\infty \gamma^k R_{t+k} | S_t=s, A_t=a]$ to evaluate a state-action pair.
Prevalent continuous deep RL algorithms \cite{lillicrap2016continuous,schulman2017proximal,fujimoto2018addressing,haarnoja2018soft} primarily follow the actor-critic architecture. In this framework, an actor $\pi_\theta$ models the policy, while a critic $Q_\phi$ learns an approximation of the Q-function conditioned by $\pi_\theta$. The actor is typically optimized by policy gradient \cite{silver2014deterministic,sutton1999policy} based on Q-value approximated by the critic.

\subsection{Policy Regularization}
Maximum entropy RL \cite{ziebart2008maximum,todorov2008general,toussaint2009robot} is a classic framework to optimize both expected return and diversity.
Haarnoja et al. introduce a deep learning framework for entropy-regularized RL \cite{haarnoja2018soft}, which modifies the standard RL objective by adding an entropy regularization $J(\pi) = \E_{\pi}[\sum\nolimits_{t=0}^\infty \gamma^t (R_t + \alpha \H(\pi(\cdot|S_t))]$, where $\H(\cdot)$ indicates the entropy of a given distribution. The state-action value function is defined as $Q^\pi_{\text{soft}}(s, a) = \E[R_t + \sum_{k=1}^\infty \gamma^k (R_{t+k} + \alpha \H(\piat{s}) | S_t=s, A_t=a]$.
The policy is optimized by applying a soft policy iteration (SPI) operator as follows
\begin{equation}
  \pinew \gets \argmin_{\pi} \mathrm{D_{KL}} \left(
    \pi(\cdot|s) \left\| \frac{\exp(Q_{\text{soft}}^{\piold} (s, \cdot))}{Z^{\piold} (s)} \right.
  \right),    
\end{equation}
where $Z^{\piold} (s)$ is a scalar to normalize the distribution, which is intractable but negligible in practice. 

Wang et al. extend the entropy-regularized RL framework to a general form \cite{wang2024negatively}. Given an arbitrary regularization function $\varrho(\pi(\cdot|s))$, and a regularized learning objective $J(\pi) = \E_{\pi}[\sum\nolimits_{t=0}^\infty \gamma^t (R_t + \alpha \varrho(\pi(\cdot|S_t))]$, the regularized Q-value is defined in the following form
\begin{equation}
    Q_{\text{reg}}^\pi(s, a) = \E_{\pi} \left[
        R_t + \left. \sum\nolimits_{k=1}^\infty \gamma^k \big( R_{t+k} + \alpha \varrho(\piat{S_{t+k}}) \big) \right| S_t=s, A_t=a  
    \right].
    \label{eq:regq}
\end{equation}
With the regularized Q-value defined, the policy can be optimized by regularized policy iteration with a policy improvement operator expressed as
\begin{equation}
    \pinew(\cdot|s) \gets \argmax_{\pi}  \left[\alpha \varrho(\piat{s}) + \E_{a \sim \pi(\cdot|s)} [Q_{\text{reg}}^{\piold}(s, a)]  
    \right]
    \label{eq:regpiimprov}
\end{equation}
for all $s \in \SS$. Furthermore, the policy gradient of the regularized learning objective is
\begin{equation}
    \nabla_\theta J(\pi) = \E_{s \sim d^{\pi}} \left[
        \alpha \nabla_\theta \varrho(\pitat{s}) + \int_\A Q^\pi_{\text{reg}}(s, a)\nabla_\theta \pi_\theta(a|s) ~\de a
    \right],
    \label{eq:regpg}
\end{equation}
where $d^\pi$ is the discounted state distribution given $\pi$.

\section{Method\label{sec:method}}

We propose diversity-regularized actor-critic (DrAC), which learn intractable multimodal policies through reparameterization and diversity regularization. Fig. \ref{fig:overview} illustrates the key ingredients of DrAC.
A detailed pseudo-code is provided in Appendix \ref{sec:code}.

\begin{figure}[h]
    \begin{subfigure}{0.44\linewidth}
        \includegraphics[width=\linewidth]{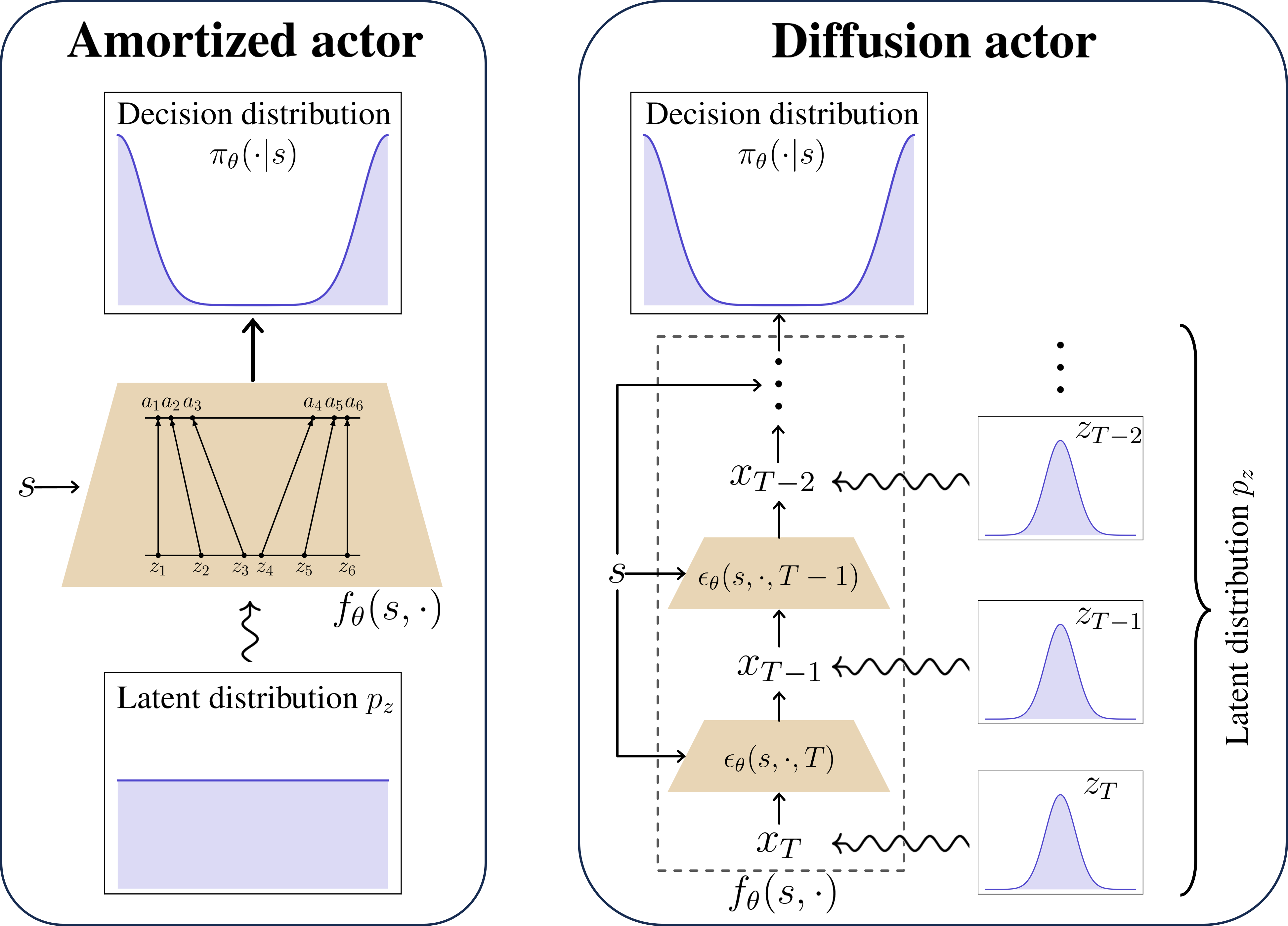}        
    \caption{We unify amortized actors and diffusion actors within a \textit{stochastic-mapping} formulation $\pi_\theta = \{f_\theta, p_z\}$ (See Section \ref{sec:policies}).}
    \end{subfigure}
    \hfill
    \begin{subfigure}{0.53\linewidth}
        \includegraphics[width=\linewidth]{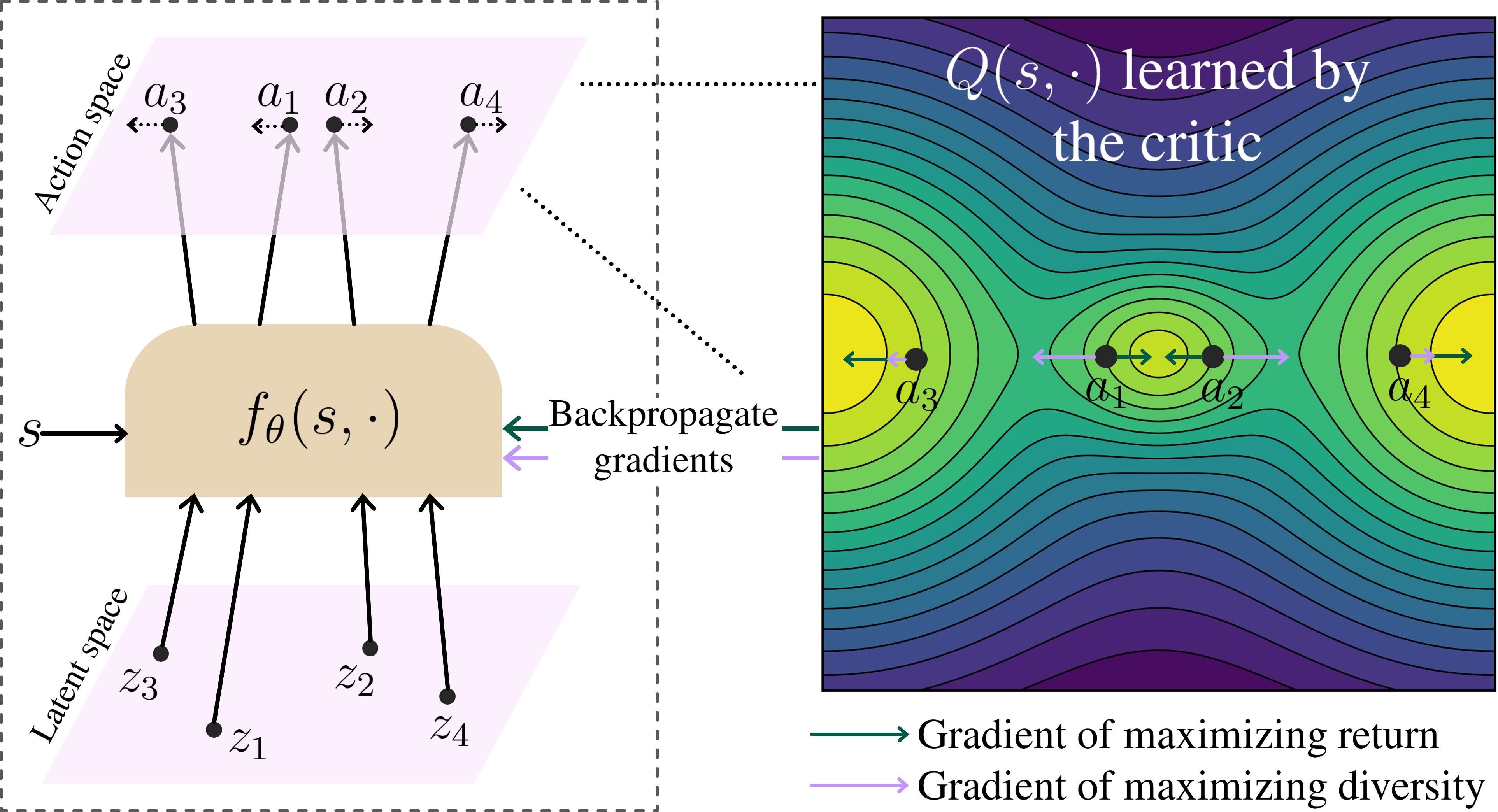}        
    \caption{We train a diversity-regularized critic and directly backpropagate the gradient of $Q$ and diversity to the actor's weights, as $\nabla_\theta Q(s,f_\theta(s,z))$ is an estimator of policy gradient via reparameterization trick.}
    \end{subfigure}
    \caption{Key ingredients of our proposed method.}
    \label{fig:overview}
\end{figure}

\subsection{Unifying Intractable Multimodal Policies and Policy Gradient\label{sec:policies}}

We define a class of \textit{stochastic-mapping actors} as $\pi_\theta = \{f_\theta, p_z\}$, which is a combination of a parameterized function $f_\theta(\cdot, \cdot): \SS \times \Z \mapsto \A$, and a fixed latent distribution $p_z$ over a latent space $\Z$. The action is drawn by sampling an independent random variable $z$ from $p_z$ first, and then feeding state $s$ and $z$ into $f_\theta$, i.e., $a \sim \pitat{s} \equiv a \gets f_\theta(s,z), z \sim p_z$. We reformulate existing multimodal actors, namely \textit{amortized actors} and \textit{diffusion actors}, under this definition as follows.
\paragraph{Amortized actors}\footnote{This definition follows \cite{haarnoja2017reinforcement} and \cite{messaoud2024s}. It is narrower than the concept of amortized inference.} This type of actors \cite{haarnoja2017reinforcement} employ a neural network (NN) $g_\theta$, which takes the state $s$ and the latent variable $z$ as input, and directly output the action $a$. We follow the most straightforward mechanism in SQL, concatenate $s$ and $z$ directly feed into $g_\theta$. Formally, this is 
\begin{align}
    f^{\text{Amort}}_\theta(s, z) \equiv g_\theta(s \oplus z).
\end{align}

\paragraph{Diffusion actors} This type of actors \cite{psenka2024learning,ren2024diffusion,wang2024diffusion} build upon the diffusion models \cite{ho2020denoising}, which are powerful generative models. We adopt the implementation from DACER \cite{wang2024diffusion}. The policy is parameterized by an NN $\epsilon_\theta(\cdot, \cdot, \cdot)$, and can be formulated by $\pi_\theta = \{f_\theta, p_z\}$ as follows
\begin{equation}
\begin{aligned}
    f^{\text{Diffus}}_\theta(s, z) &\equiv x_0,  \\
    x_{t-1} &= \frac{1}{\sqrt{\alpha_t}} \left( 
        x_t - \frac{\beta_t}{\sqrt{1 - \bar \alpha_t}} \epsilon_\theta(s, x_t, t)
    \right)  + \sqrt{\beta_t} z_{t-1}, \\
    x_T &= z_T.
\end{aligned}    
\label{eq:diffus}
\end{equation}
In Eq. \eqref{eq:diffus}, $\alpha_t$ and $\beta_t$ are scalars calculated by some specific schedule mechanisms to control the diffusion process, $z_0, z_1, \cdots, z_T$ are i.i.d. noise vectors drawn from $\gaus(\mathbf{0}, I)$ and we have treated $z = \{z_0, \cdots, z_T\}$, $T$ is the number of diffusion steps. Note that the $t$'s here are not the time steps of MDP but the time steps of the diffusion process.

These actors are powerful, but their decision probabilities $\pi_\theta(a|s)$ does not have a closed-form expression, making policy gradient not directly applicable. However, as they follow the stochastic mapping formulation, it is feasible to calculate policy gradient via reparameterization trick (PGRT) as mentioned in \cite{lan2022model} by the equation below
\begin{equation}
    \label{eq:pgrt}
    \nabla_\theta J(\pi_\theta) = \E_{s \sim d^\pi, z \sim p_z} [\nabla_a Q(s,f_\theta(s,z)) \nabla_\theta f_\theta(s,z)].
\end{equation}

PGRT demonstrates that we can directly backpropagate the Q-function's gradient to $f_\theta$ to train stochastic-mapping actors, where the Q-function can be approximated by differentiable critics. We noticed that DACER \cite{wang2024diffusion} backpropagates the gradient of the Q-function to train diffusion actors, but the paper of DACER did not mention the relation to policy gradient and did not disclose that this method can be generalized to other actors. 

\subsection{Actor-Critic Learning with Diversity Regularization}

As we pursue decision diversity, an effective diversity regularization is desired. As discussed earlier, the traditional entropy regularization do not prefer multimodality and is not applicable to intractable actors. To overcome this challenge, we propose a distance-based regularization to encourage multimodal diversity. 

\begin{wrapfigure}{r}{0.35\linewidth}
    \vspace{-1.2em}
    \includegraphics[width=\linewidth]{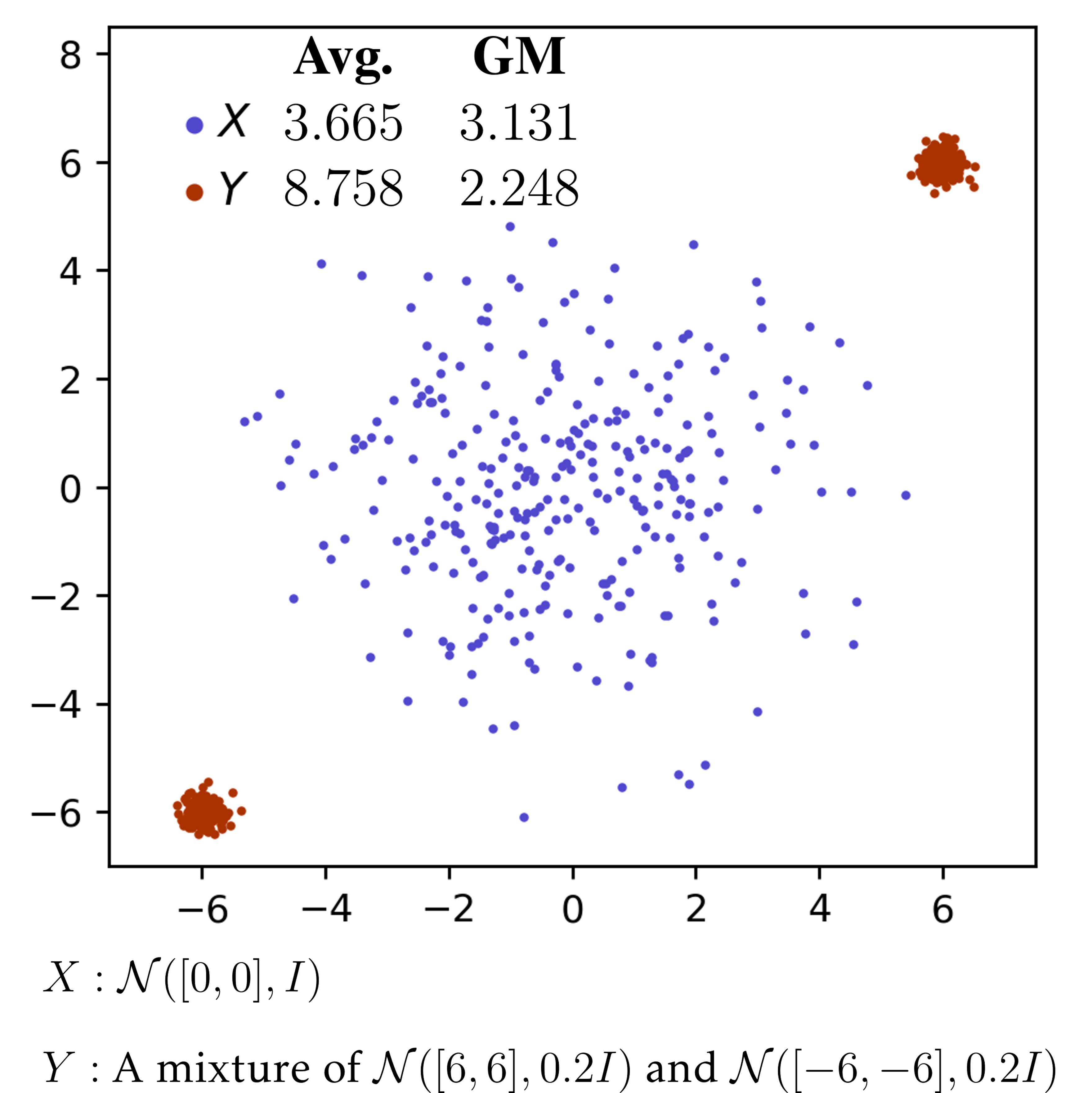}
    \caption{Average (Avg.) and geometric mean (GM) of pairwise distances for two synthesized data distributions. Average pairwise distance overestimates the diversity of distribution $Y$.}
    \label{fig:gm-motivation}
    \vspace{-0.5em}
\end{wrapfigure}
A straightforward distance-based diversity metric is the mean pairwise distance, formulated as 
$D^\pi(s) = \E_{x, y \sim \piat{s}}[\delta(x, y)]$,
where $\delta(\cdot, \cdot)$ is a distance metric.
However, this metric may overestimate the diversity when the data distribution forms some tiny clusters far away from each other (See Fig. \ref{fig:gm-motivation}). We empirically found that this issue led to underwhelming performance.
Therefore, we consider the geometric mean $\gm(\{\delta(x_1, y_1), \cdots, \delta(x_n, y_n) \}) = \left( \prod\nolimits_{i=1}^n \delta (x_i, y_i) \right)^{\nicefrac{-1}{n}}$ of pairwise distances instead. The geometric mean is sensitive to small values as multiplication is taken instead of summation, which mitigates the overestimation issue (See Fig. \ref{fig:gm-motivation}).
Furthermore, we reshape diversity on the log scale to make it easier to balance reward and diversity. We propose a diversity regularization as follows
\begin{equation}
    D^\pi(s) = \E_{x \sim \piat{s}, y \sim \piat{s}}[\log\delta(x, y)].
\end{equation}
Specifically, L2 distance is adopted as the distance metric. This diversity regularization is added to the original RL objective with a coefficient $\alpha$, resulting in an objective to maximize as
\begin{equation}
    \label{eq:overallJ}
    J = \E_{\pi} \left[\sum\nolimits_{t=0}^\infty \gamma^t (R_t + \alpha D^\pi(S_t)) \right].
\end{equation}
We estimate the diversity of the actor $\pi_\theta$ at a certain state $s$ by sampling $n$ pairs of i.i.d. latent vectors $(z^x_1, z^y_1), \cdots, (z^x_n, z^y_n)$, and compute the average log-distance over samples. This is denoted as $\tilde D_\theta(s) = \frac{1}{n} \sum_{i=1}^n \log\delta(f_\theta(s, z^x_i), f_\theta(s, z^y_i))$. Though this estimation requires multiple samples, it is time-efficient as such a process can be easily parallelized on GPUs.

The practical DrAC algorithms follow the common actor-critic architecture. The actor is a stochastic-mapping actor, and the critics follow the learning tricks in soft actor-critic (SAC) \cite{haarnoja2018soft}. Double critics $Q(\cdot, \cdot; \phi_1), Q(\cdot, \cdot; \phi_2)$ and double target critics $\hat Q(\cdot, \cdot; \hat\phi_1), \hat Q(\cdot, \cdot; \hat\phi_2)$ are employed.
By substituting $\tilde D_\theta$ into Eq. \eqref{eq:regq} and unrolling it into a bootstrap learning target, the critic loss is derived as
\begin{equation}
    \L_\phi = \E_{s, a, r, s' \sim \mathcal{D}} \left[\mse (Q(s, a; \phi_i), r + \gamma (\tilde V(s'; \hat \phi) + \alpha \tilde D_\theta(s')) \right], i \in \{1, 2\}
\end{equation}
where $\mathcal{D}$ represents the off-policy replay buffer, $\tilde V(s'; \hat \phi) = \min_{i \in \{1, 2\}} \hat Q(s', a'; \hat\phi_i)$ is an approximation of the value of the next state $s'$ based on the target critics and $a'$ is a sample drawn from $\pitat{s'}$. The target critics are updated by the target smoothing strategy \cite{haarnoja2018soft,fujimoto2018addressing}.

Regarding actor learning, we combine PGRT (See Eq. \eqref{eq:pgrt}) with the regularized policy gradient theorem (See Eq. \eqref{eq:regpg}), derive a loss function as follows
\begin{equation}
    \L_\theta = -\E_{s \sim \mathcal{D}, z \sim p_z} [Q(s, f_\theta(s, z); \phi) + \alpha \tilde D_\theta(s)],
\end{equation}
where $Q(s, f_\theta(s, z); \phi) = \min_{i \in \{1, 2\}} Q(s, f_\theta(s, z); \phi_i)$, following the technique used in SAC.
This loss function can also be considered an approximation of Eq. \eqref{eq:regpiimprov}. 

\subsection{Automatic Coefficient Adjustment based on Target-Diversity}
Multimodal actors are expressive and flexible. However, given the diversity regularization with a fixed coefficient, such properties can also lead to instability in critic learning, which may harm the training.
To mitigate this issue and also make hyperparameter tuning easier, we borrow the automatic coefficient adjustment technique from SAC \cite{haarnoja2018sac}. 
Specifically, we parameterize $\alpha$ with a learnable scalar $w$ such that $\alpha = \exp(w)$ if $w < 0$ and $\alpha = w + 1$ otherwise.
Then $\alpha$ is treated as learnable and updated by descending the gradient of a dual-optimization loss defined as
\begin{equation}
    \L_\alpha = \E_{s \sim \mathcal{D}} [\alpha(\tilde D_\theta(s) - \hat D)],
\end{equation}
where $\hat D$ is a target diversity value. Minimizing this loss will make $\tilde D_\theta(s)$ match with $\hat D$ adaptively.
As $\tilde D_\theta(s)$ is defined on the log-scale, determining its range can be trivial. We introduce a temperature hyperparameter $\beta$ such that $\hat D \gets \log(\beta\sqrt{|\A|})$ to ease the triviality. The scale of $\beta$ is comparable to pairwise distances after normalization by the action space dimensionality $|\A|$.

\section{Experiments\label{sec:experiments}}

\paragraph{Domains}
Previous works on multimodal RL \cite{haarnoja2017reinforcement,messaoud2024s,wang2024diffusion} only verify the multimodal ability with a toy 2D multi-goal navigation task and mainly focus on the improvements in MuJoCo. To demonstrate the advantages of DrAC and multimodal policies, we include two diversity-critical benchmarks in \textit{multi-goal achieving} and \textit{generative RL} domains.
We also test DrAC and representative multimodal RL algorithms in standard MuJoCo benchmarks to compare the general performance.

\paragraph{Baselines}
By installing DrAC with an amortized actor and a diffusion actor separately, we obtain two versions of DrAC, namely DrAmort and DrDiffus. We include SQL \cite{haarnoja2017reinforcement}, DACER \cite{wang2024diffusion}, S$^2$AC \cite{messaoud2024s} as multimodal RL baselines, and SAC \cite{haarnoja2018sac} as a unimodal baseline. Table \ref{tab:algos} summarizes the techniques of these algorithms. We keep NN architectures and common hyperparameters all the same as the default setting of SAC to enable a fair comparison \cite{haarnoja2018soft}. Appendix \ref{sec:hyperparams} details the hyperparameters. All trainings are conducted with five seeds.

\begin{table}[h]
    \centering
    \caption{A summary of compared algorithms. Bold texts indicate our contributions. The ``Temperature'' column indicates the technique of adjusting the emphasis on diversity used by each algorithm.}
    \resizebox{\linewidth}{!}{
        \begin{tabular}{l|c|c|c|c|c}
   \toprule
      & Actor Model & Actor Learning & Critic Learning & Diversification & Temperature  \\
   \midrule
     \textbf{DrAmort} & Amortized & \textbf{Regularized PGRT} & Conventional & \textbf{Max-Diversity} & Target-Matching \\
     SQL \cite{haarnoja2017reinforcement} & Amortized & SPI via SVGD & Conventional & Max-Entropy & Constant \\
     \textbf{DrDiffus} & Diffusion & \textbf{Regularized PGRT} & Conventional & \textbf{Max-Diversity} & Target-Matching \\
     DACER \cite{wang2024diffusion} & Diffusion & PGRT & Distributional \cite{duan2025distributional} & Noise Scaling & Target-Matching \\
     S$^2$AC \cite{messaoud2024s} & Energy-based & SPI via SVGD & Conventional & Max-Entropy & Constant \\
     SAC \cite{haarnoja2018soft} & Gaussian & SPI & Conventional & Max-Entropy & Target-Matching \\
   \bottomrule
\end{tabular}

    }
    \label{tab:algos}
\end{table}

\subsection{Multi-goal Achieving}
In real-world scenarios, there could be multiple goals to achieve. We build a multi-goal version of the PointMaze environment in D4RL \cite{fu2020d4rl} to conduct a case study of multi-goal achieving. 
This environment requires the agent to learn to steer a ball from a fixed original point to navigate to one of the goals within a maze. The maze contains multiple goals. The reward is sparse, only reaching a goal returns $+100$ reward and ends the episode with a success flag.
Besides the success rate, the number of reachable goals is also evaluated. We designed three maze maps with increasing difficulty levels, as illustrated in Fig. \ref{fig:pm-maps}.
We notice that the work of \cite{li2024learning} tested more complex multi-goal achieving benchmarks. However, that work autonomously collects and clusters trajectories first, then clones behaviors using these clusters, while our work focuses on basic RL algorithms. 
To ensure a fair and meaningful comparison, we conducted a grid search for each algorithm before the formal evaluation. The highest temperature that ensures each algorithm reaches the optimal success rate is picked.

\begin{figure}[h]
    \centering
    \includegraphics[width=\linewidth]{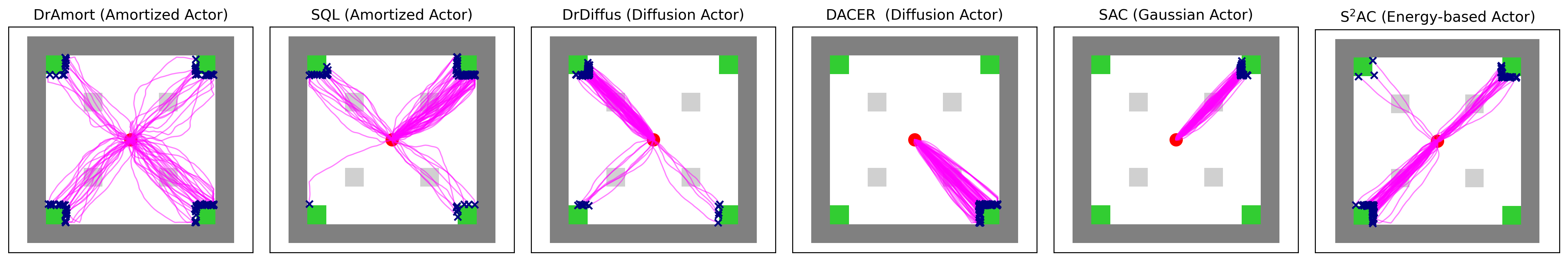}
    \caption{Evaluation trajectories of all tested algorithms in the simple maze.}   
    \label{fig:trajs}
\end{figure}

\begin{figure}[h]
    \begin{subfigure}{0.49\linewidth}
    \includegraphics[width=\linewidth,trim=3 3 3 6,clip]{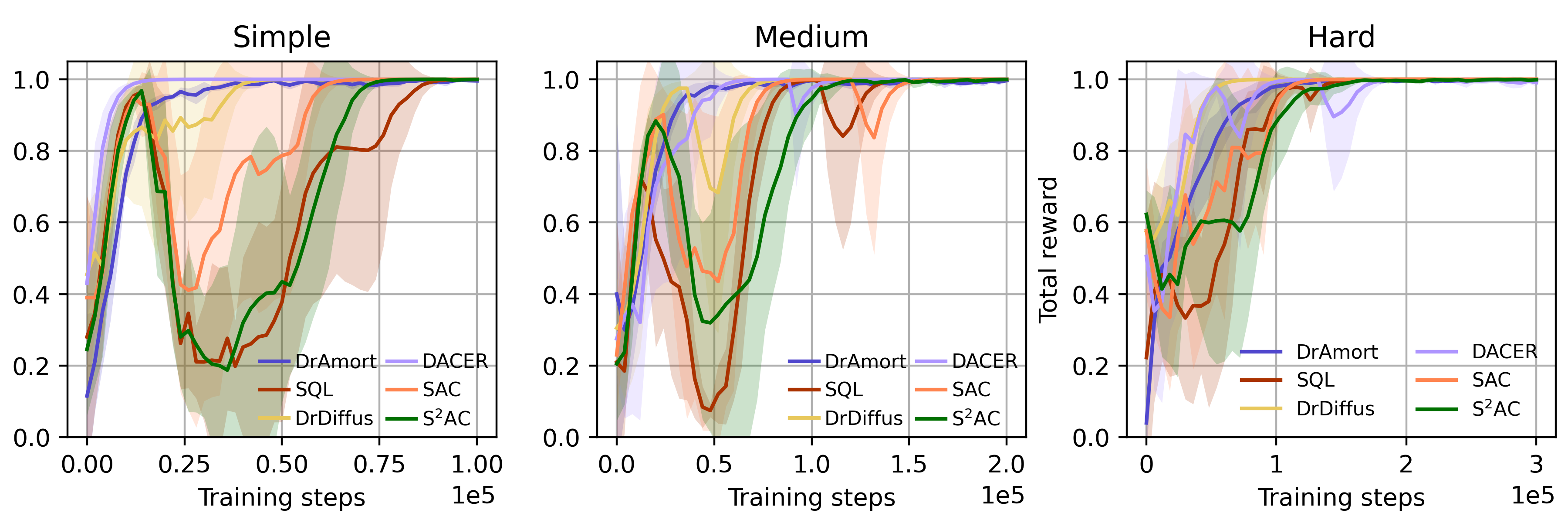}
    \caption{Success rate}   
    \label{fig:pm-sr}
    \end{subfigure}
    \hfill
    \begin{subfigure}{0.49\linewidth}
    \includegraphics[width=\linewidth,trim=3 3 3 6,clip]{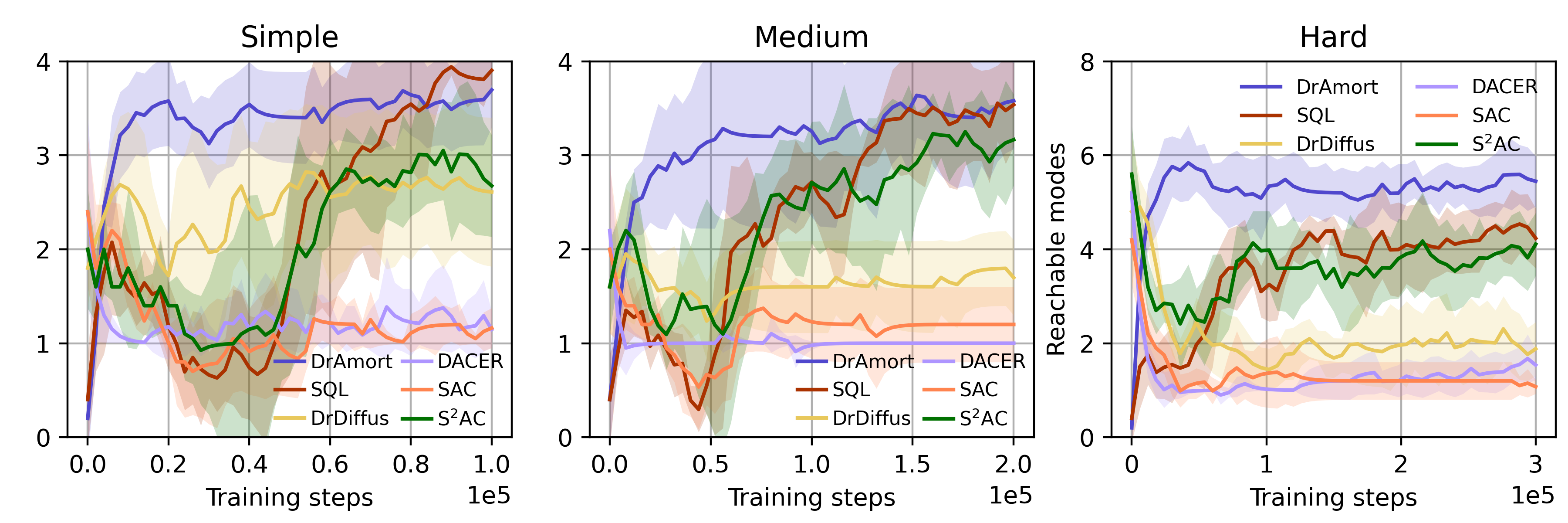}
    \caption{Number of reachable goals}        
    \label{fig:pm-modes}
    \end{subfigure}  \\
    \begin{subfigure}{0.49\linewidth}
    \includegraphics[width=\linewidth,trim=3 3 3 6,clip]{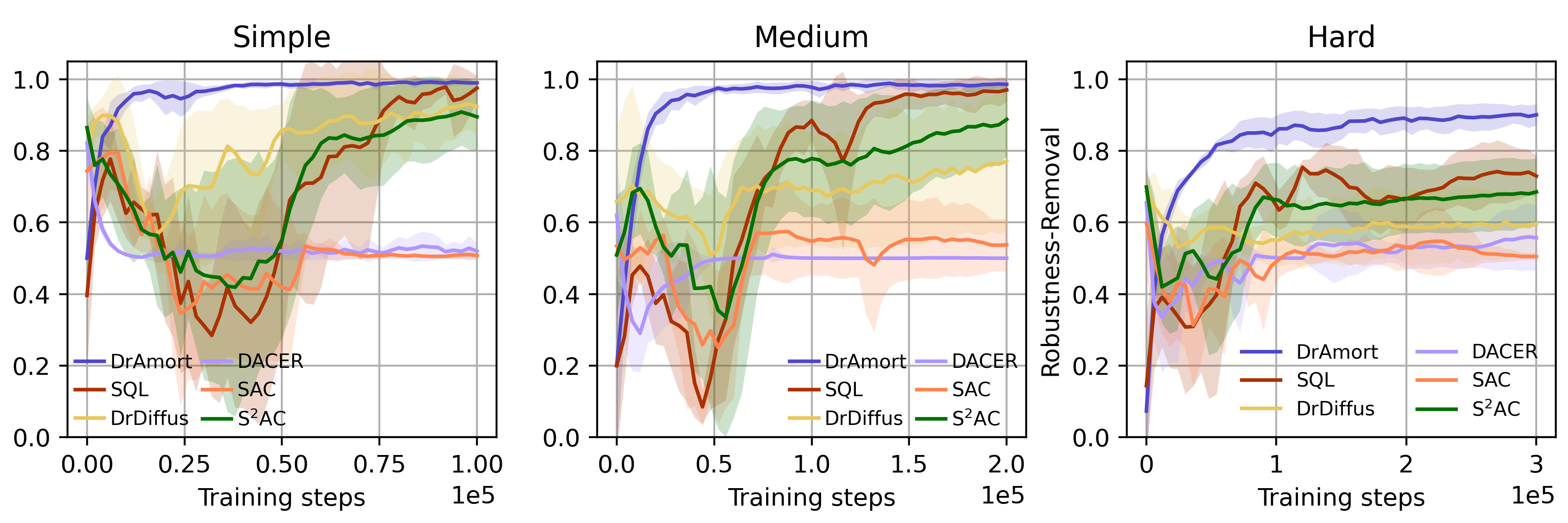}
    \caption{Robustness against \textit{removal}}   
    \label{fig:pm-removal}
    \end{subfigure}
    \hfill
    \begin{subfigure}{0.49\linewidth}
    \includegraphics[width=\linewidth,trim=3 3 3 6,clip]{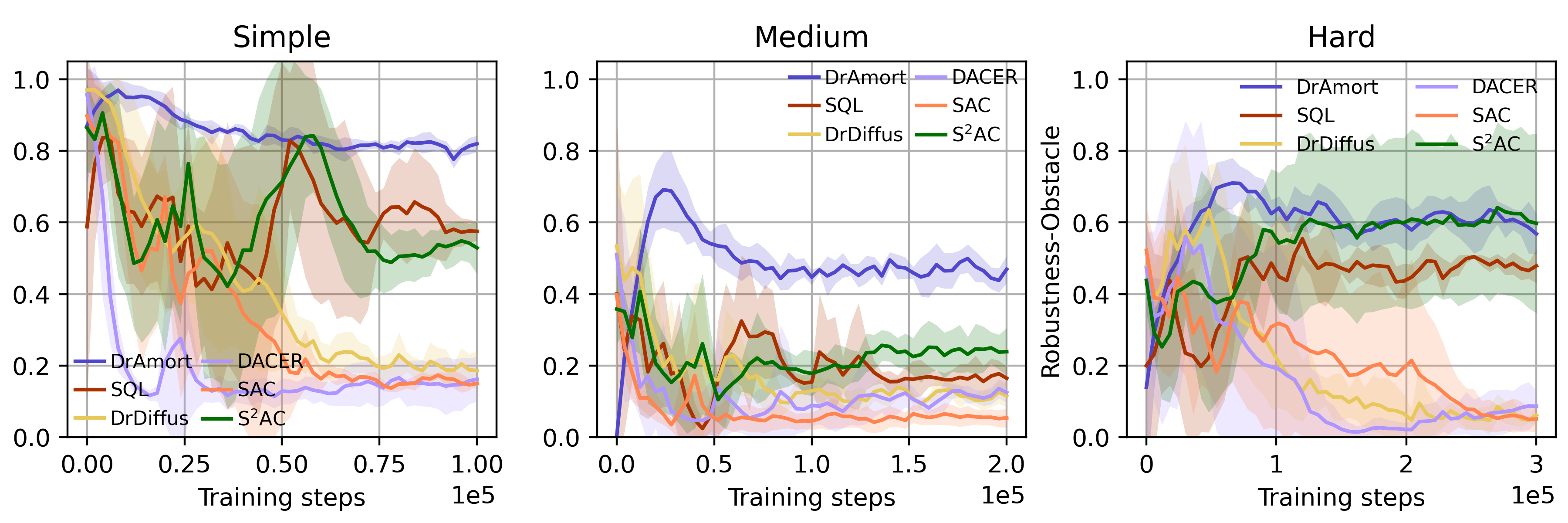}
    \caption{Robustness against \textit{obstacle}}        
    \label{fig:pm-obstacle}
    \end{subfigure}
    \caption{Learning curves in multi-goal PointMaze. Solid lines represent mean performance, and shaded regions indicate standard deviation. All curves are smoothed by the exponential moving average with a coefficient of $0.5$.}
    \label{fig:pm-curves}
\end{figure}

The trajectories in the simple maze of learned policies are visualized in Fig. \ref{fig:trajs}. It is clear that DrAmort learns the most diverse and uniform trajectories. Despite using the same diffusion actor, DrDiffus learn to reach different goals, DACER does not. Fig. \ref{fig:pm-modes} shows that DrAmort and SQL consistently learns to reach the most goals in all mazes, and DrAmort learns faster than SQL. In terms of success rate, DrAmort, DrDiffus, and DACER learn fast and stably, while SQL, SAC, and S$^2$AC exhibit a decrease in success rate in the early stage and reach optimal success rate more slowly. Given the results, we consider that amortized actors possess superior multimodal expressivity. DACER struggles with learning multimodal behaviors in PointMaze. This is most likely due to DACER only controlling the diversity via noise scaling. Meanwhile, SQL shows better multimodality than DrDiffus. The underwhelming multimodality of diffusion actors may be owing to a small number of diffusion steps of $20$ we used as suggested in \cite{wang2024diffusion}. However, despite using a small number of diffusion steps, the inference and training speed is much slower than amortized actors. Therefore, we advocate that the amortized actor is a promising actor class for representing multimodal policies. Trajectories in medium and hard mazes are presented in Appendix \ref{sec:moretrajs}.
 
\vspace{-5pt}

Inspired by \cite{kumar2020one}, we investigate whether diversity empowers few-shot robustness in out-of-distribution scenarios. We consider two types of perturbations: 1) \textit{Removal}: half of the goals are removed during the test; 2) \textit{Obstacle}: some obstacles that do not exist in the training phase, appear in the test phase and block the shortest paths to the goals. 
The first case requires agents to reach all goals with uniform frequency. The second case requires the agents to navigate to the goal with diverse trajectories, so that the they can bypass the obstacles by chance. Similar to \cite{kumar2020one}, we test the \textit{five-episode success rate} of all algorithms in each scenario. That is, the probability that the agent can successfully reach any goal at least once within five independent trials. 
The results are shown in Fig. \ref{fig:pm-removal}  and Fig. \ref{fig:pm-obstacle}. 

DrAmort shows the best robustness in all mazes, while SQL outperforms others among the rest algorithms.  
According to Fig. \ref{fig:trajs}, DrAmort exhibits the best trajectory diversity, so it has more chances to bypass the obstacles. Meanwhile, DrAmort reaches goals more uniformly, so higher robustness against removal is guaranteed.
The results confirm that multimodality and diversity empower few-shot robustness in out-of-distribution scenarios. DrDiffus outperforms DACER regarding robustness against removal. 
Regarding SAC, although a high temperature is set, it struggles in learning to reach multiple goals and generalizing to out-of-distribution test cases. This phenomenon confirms the fundamental limitation of unimodal actors.

\subsection{Generative RL}
There has been some application of RL in generative tasks, e.g., game content generation \cite{khalifa2020pcgrl,wang2024negatively} and program generation \cite{liu2023hierarchical}. We take game level generation benchmark in \cite{wang2024negatively} as a case study because policy multimodality is important to balance quality and diversity in this domain, according to \cite{wang2024negatively}.

\begin{figure}[h]
    \includegraphics[width=0.495\linewidth]{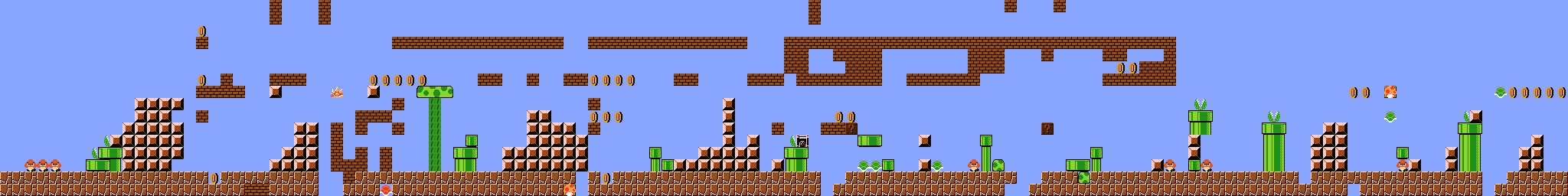} 
    \hfill
    \includegraphics[width=0.495\linewidth]{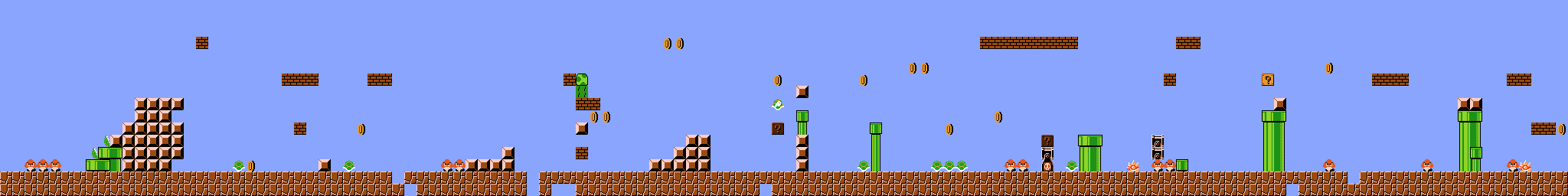}
    \caption{Example \textit{MarioPuzzle} style (left) and \textit{MultiFacet} style (right) levels generated by DrAmort.}
    \label{fig:smblvls}
\end{figure}

In this game level generation benchmark, the agent needs to observe a sliding window on an existing game level and propose new level segments to be concatenated with the existing level. The level segment is represented by a decoder, which is the generator of pretrained generative adversarial networks \cite{goodfellow2014generative,volz2018evolving}. That means the agent will output a continuous latent vector at each step, and the decoder will decode the action into a piece of level. There are two expert knowledge-based formulations of the reward function in this benchmark, which aim to evaluate the quality of generated levels under some game design principles. The reward also includes a penalty for unsolvable level segments to guide agents to generate solvable levels. Solvable or not is checked by rules in our experiments.
Different reward functions result in different styles of generated levels, namely \textit{MarioPuzzle} and \textit{MultiFacet}. Fig. \ref{fig:smblvls} shows example levels of the two styles generated by DrAmort. 
Besides episodic return, the diversity of generated levels is also desired in this domain.

\begin{figure}[h]
    \begin{subfigure}{0.49\linewidth}
    \includegraphics[width=\linewidth,trim=6 6 6 6,clip]{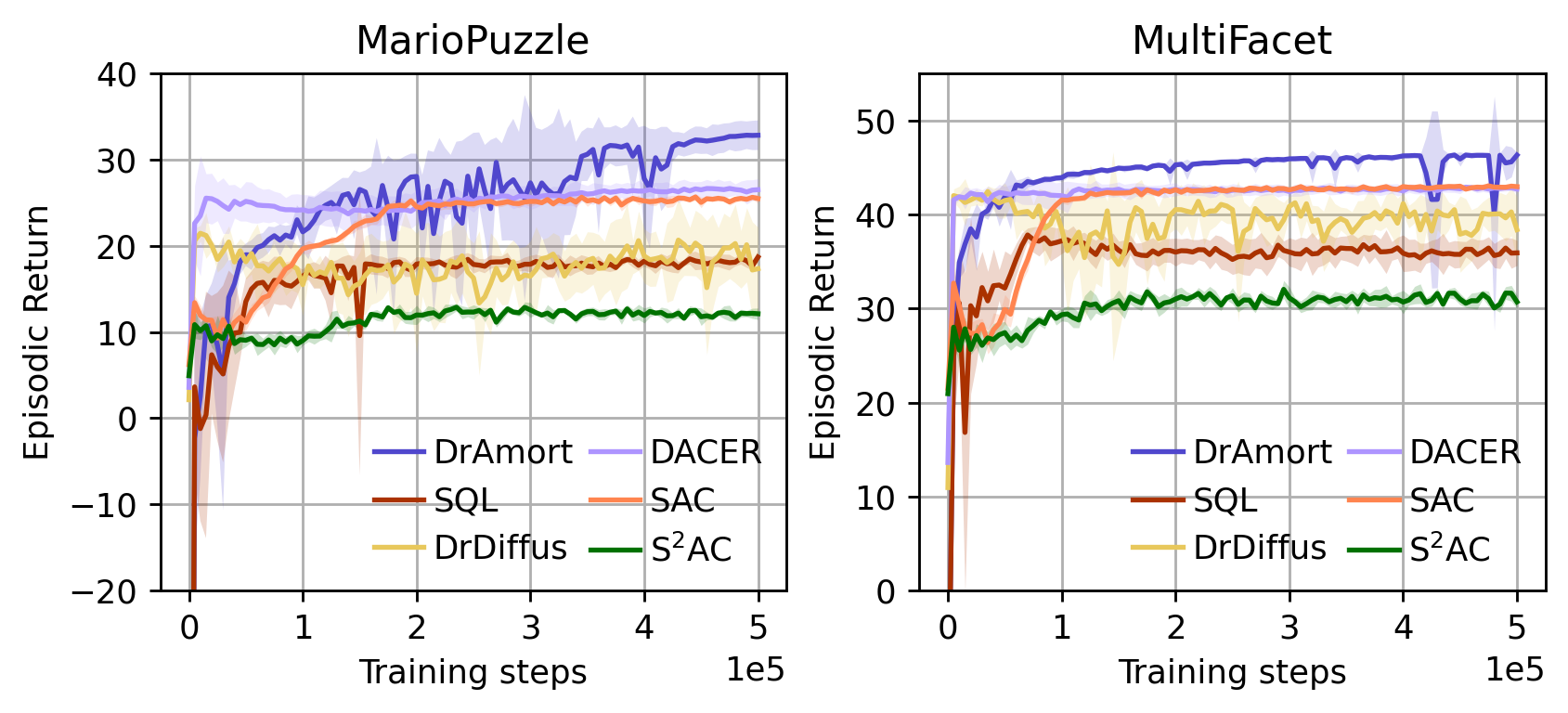}    
    \caption{Learning curves regarding average episodic return, i.e., average quality of generated levels.}
    \label{fig:smbgen-curves}
    \end{subfigure}
    \hfill
    \begin{subfigure}{0.49\linewidth}
    \includegraphics[width=\linewidth,trim=6 6 6 6,clip]{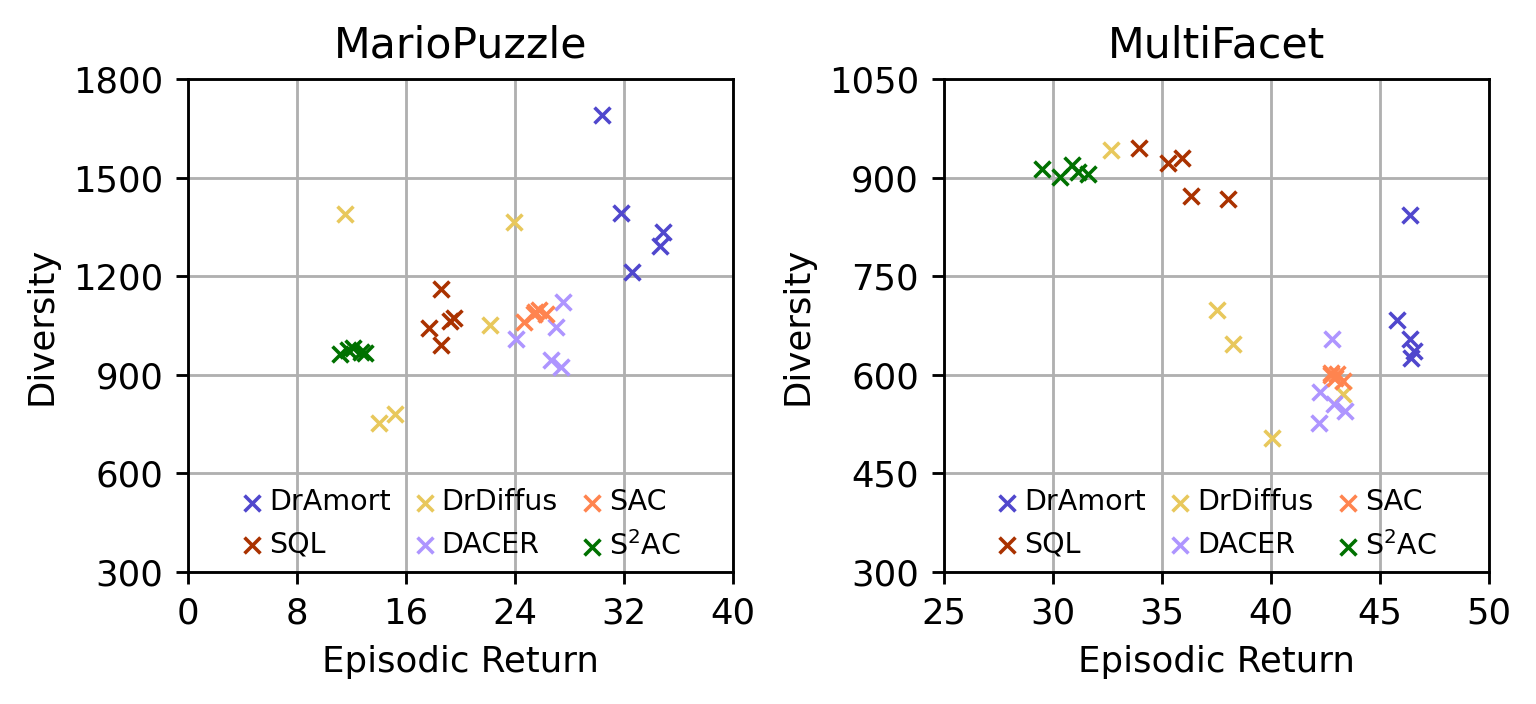}                
    \caption{Locations of all trained policies in the episodic return-diversity performance space.}
    \label{fig:smbgen-qd}
    \end{subfigure}
    \caption{Evaluation results in the game content generation benchmark.}
    \label{fig:smbgen}
\end{figure}

We train algorithms with the same temperature settings as multi-goal PointMaze. The learning curves are shown in Fig. \ref{fig:smbgen-curves}. For both styles, DrAmort shows the best episodic returns, namely the best qualities. This is possibly attributed to the superior expressivity of the amortized actor. As the distribution of high-quality content is most likely multimodal, only highly expressive actors can represent high-performance policies under high temperatures. Though SQL also applies amortized actor, it performs badly.
As SQL directly uses SVGD as the gradient estimator for approximating SPI, it is possible that reparameterization trick is a better gradient estimator than SVGD for RL. S$^2$AC, which applies SVGD to directly sample actions, performs the worst on all tasks. On the other hand, DrDiffus and DACER do not outperform SAC.

We also evaluate the diversity of generated levels by computing the average pairwise Hamming distance following \cite{wang2024negatively}, and then visualize the performance in the 2D quality-diversity objective space of all trained policies in Fig. \ref{fig:smbgen-qd}. Each marker in the figure represents a policy trained by one of the algorithms with an independent seed. Under MarioPuzzle style, DrAmort generally dominates other algorithms, i.e., outperforms other algorithms in terms of both return and diversity. Under the MultiFacet style, DrAmort dominates other algorithms except for SQL. Though SQL shows the best diversity, its episodic return is poor. 
Under the other style, SQL does not demonstrate better diversity, and the episodic return is still poor.
DrDiffus, DACER, and SAC are non-dominated with each other. But diffusion actor is much slower than Gaussian and amortized actors. Therefore, diffusion actors might not be a good choice for this task. 
We present t-SNE embedding visualizations \cite{van2008visualizing} of generated levels in Appendix \ref{sec:tsne}, to demonstrate their diversity more intuitively.

\subsection{Conventional Benchmarks}
\begin{wrapfigure}{r}{0.6\linewidth}
    \vspace{-1.25em}
    \includegraphics[width=\linewidth,trim=6 6 6 6,clip]{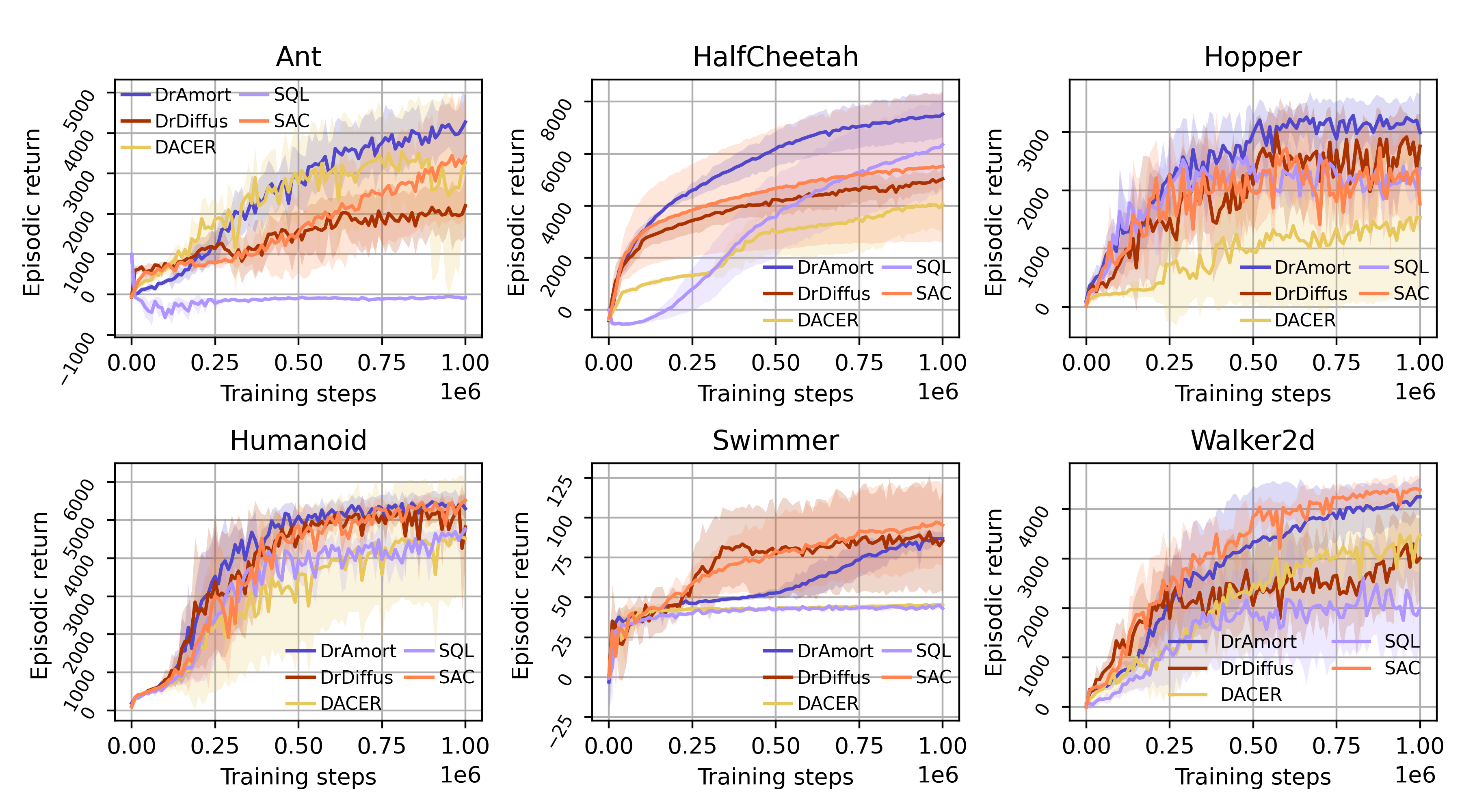}
    \caption{Learning curves in MuJoCo.}
    \label{fig:mujoco}
\end{wrapfigure}

To evaluate the general performance of DrAC in conventional tasks, we test DrAC in MuJoCo-v4 and compare it with baselines. We do not include S$^2$AC as it is too computationally intensive and it uses an unusual implicit actor, making it less meaningful to compare standard performance. 
Learning curves in the six MuJoCo locomotion tasks are presented in Fig. \ref{fig:mujoco}. 

DrAmort exhibits competitive performance. It achieves the best performance in three out of the six tasks and surpasses SQL in a total of five, demonstrating that DrAC is a strong method for training amortized actors.
Overall, the performance of DrDiffus is comparable to DACER, though DACER leverages a more advanced critic learning technique \cite{duan2025distributional}. Therefore, DrAC can serve as a high-performance base algorithm for learning multimodal policies, particularly for amortized actors. Meanwhile, it is surprising that DACER and DrDiffus do not outperform SAC, though they apply more complex diffusion actors. We notice that the paper of DACER \cite{wang2024diffusion} uses deeper networks, more advanced activations, and smaller learning rates than the default settings of \cite{haarnoja2018soft}. 
It is possible that diffusion actors need more powerful networks and more elaborate hyperparameter tuning to fulfill their potentials.


\section{Conclusion\label{sec:conclusion}}
We propose \textit{\underline{D}iversity-\underline{r}egularized \underline{A}ctor \underline{C}ritic} (DrAC), a novel and versatile RL algorithm designed to train intractable multimodal policies. DrAC addresses the intractability challenge in training multimodal policies by leveraging the reparameterization trick and distance-based diversity regularization. Meanwhile, DrAC is compatible with all policy models that can be formulated as a stochastic-mapping actor defined as $\pi_\theta = \{f_\theta, p_z\}$ such that $a \sim \pi_\theta(\cdot|s) \equiv a \gets f_\theta(s,z), z \sim p_z$. Experiments show DrAC can train amortized actors to achieve superior performance in multi-goal achieving and generative RL benchmark, which also verifies the significance of multimodal policies in these domains.
Notably, using an amortized actor, DrAC demonstrates strong few-shot robustness \cite{kumar2020one} in out-of-distribution test scenarios within the multi-goal PointMaze environment. Nevertheless, this paper only investigates a basic implementation of DrAC, leaving ample room for further improvements. Future work could investigate scheduling the temperature, designing better stochastic-mapping actors, exploiting other distance and diversity metrics, and integrating more learning tricks, etc.

Our empirical results also indicate that the amortized actor possesses strong multimodal expressivity and performs well in all domains involved in this paper. On the other hand, the diffusion actor does not show significant advantages. It is possible that diffusion actors need more careful tuning of hyperparameters and better NN architecture design. However, the training and inference speed of amortized actors is comparable to traditional actors and significantly faster than diffusion actors. Therefore, we posit that the amortized actor represents a promising actor class for multimodal RL.

\begin{ack}
This work is supported by the National Natural Science Foundation of China 62406266.
\end{ack}

\bibliographystyle{plainnat}  
\small
\bibliography{main}

\newpage
\appendix

\section{Technical Details \label{sec:details}}

\subsection{Pseudo code of DrAC \label{sec:code}}
Algorithm \ref{algo:drac} details DrAC.

\begin{algorithm}[H]
    \caption{Diversity-regularized Actor-Critic (DrAC)}
    \label{algo:drac}
    \begin{algorithmic}
      \REQUIRE Actor weights $\theta$; Critic weights $\phi = \{\phi_1, \phi_2\}$; Target critic weights $\hat \phi = \{\hat\phi_1, \hat\phi_2\}$
      \STATE Let $\mathrm{gradient\_step\_count} \gets 0$
      \FOR {$t: 1 \to T$}
        \STATE Sample an action $a \sim \pi(\cdot|s)$
        \STATE Execute $a$ and observe $s', r, d$
        \STATE $r, s', d \gets$ Execute $a$ in the environment 
        \STATE Store $(s, a, r, s, d)$ in the buffer
        \WHILE{$\mathrm{gradient\_step\_count} < t * \mathrm{reply\_ratio}$ }
            \STATE Sample a batch $(s, a, r, s', d)$ from the replay memory
            \STATE Sample $z$; $(z^x_1, z^y_1), \cdots, (z^x_n, z^y_n)$ from $p_z$ 
            \STATE Let $\tilde D_\theta(s') \gets \frac{1}{n} \sum_{i=1}^n \log \delta(f_\theta(s', z^x_i), f_\theta(s', z^y_i) )$
            \STATE Compute Q-target $y \gets r + \gamma (1-d) (\hat Q(s, f_\theta(s, z); \hat\phi) + \alpha \tilde D_\theta(s'))$
            \STATE Update $\phi_1$ and $\phi_2$ with $\mse(Q(s, a; \phi_i), y)$, $i \in \{1, 2\}$
            \STATE Sample $z$; $(z^x_1, z^y_1), \cdots, (z^x_n, z^y_n)$ from $p_z$ 
            \STATE Let $\tilde D_\theta(s) \gets \frac{1}{n} \sum_{i=1}^n \log \delta(f_\theta(s, z^x_i), f_\theta(s, z^y_i))$
            \STATE Update $\theta$ with $\L_\theta = -Q(s,f_\theta(s, z); \phi) - \alpha \tilde D_\theta(s))$
            \STATE Update $\alpha$ with $\L_\alpha = \alpha(\tilde D_\theta(s) - \hat D)$
            \STATE Update $\hat \phi$ by $\hat\phi \gets \rho \phi + (1-\rho) \hat \phi$
            \STATE Let $\mathrm{gradient\_step\_count} \gets \mathrm{gradient\_step\_count} + 1$
        \ENDWHILE
      \ENDFOR        
\end{algorithmic}
\end{algorithm}

\subsection{Experiment Settings \label{sec:hyperparams}}
\begin{figure}[H]
    \centering
    \includegraphics[width=\linewidth]{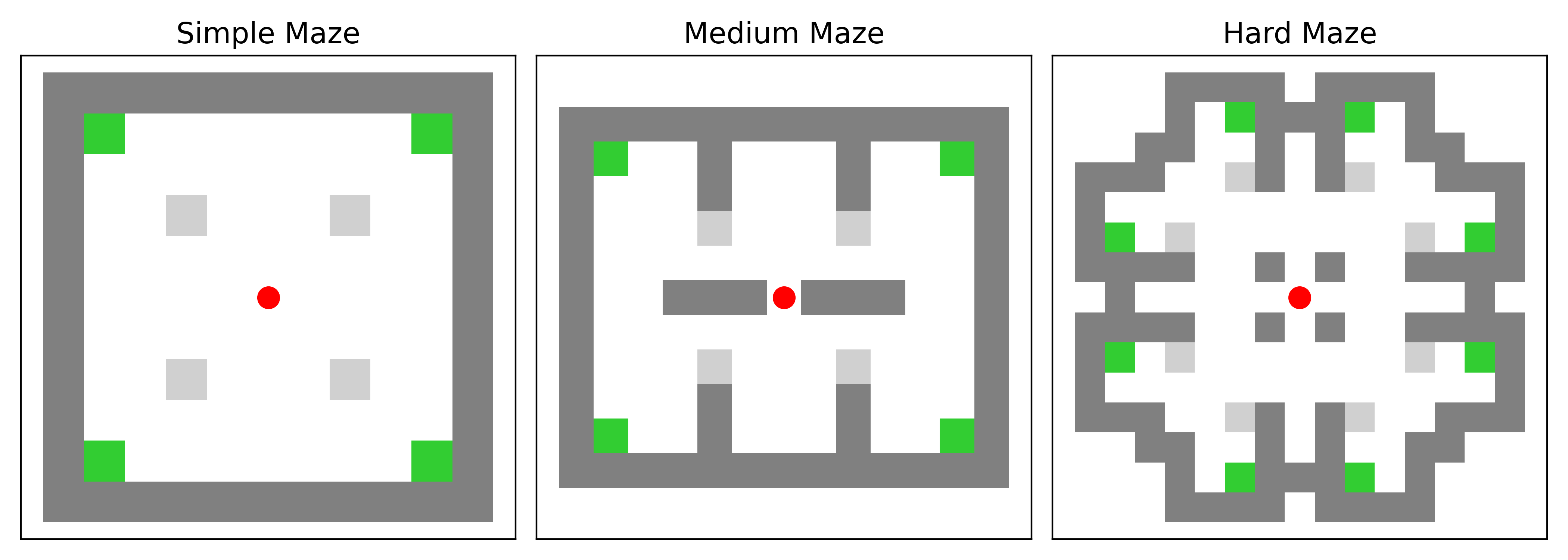}
    \captionsetup{font=small}
    \caption{Maze maps of the Multi-goal PointMaze environment. The red circle indicates the start point, green cells indicate goals, gray cells are walls, and light gray cells are obstacles that only appear in the robustness evaluation stage.}
    \label{fig:pm-maps}
\end{figure}

The customized maze maps used in our multi-goal PointMaze environment are shown in Fig. \ref{fig:pm-maps}

The hyperparameters are listed in Table \ref{tab:sharedparam}. The temperature hyperparameters listed in Table \ref{tab:sharedparam} are used in the multi-goal PointMaze environment and the game content generation environment. While for MuJoCo, we use a lower temperature for DrAC. For baseline algorithms, we use the default temperatures MuJoCo provided in their original papers. Temperature hyperparameters for all algorithms in MuJoCo are listed in Table \ref{tab:mujoco-temps}. 

\newpage
\begin{table}[t]
    \renewcommand\arraystretch{1.28}
    \caption{Hyperparameters}
    \small
    \resizebox{\linewidth}{!}{
    \begin{tabular}{l|c|c|c|c|c|c}
    \toprule
     Hyperparameter & \textbf{DrAmort} & \textbf{DrDiffus} & DACER & SQL & S$^2$AC & SAC \\
    \midrule
     Replay buffer size & \multicolumn{6}{c}{1000000} \\
     MLP hidden layers in actors \& critics & \multicolumn{6}{c}{(256, 256)}  \\
     Hidden layer activation in all networks & \multicolumn{6}{c}{ReLU} \\
     Optimizer & \multicolumn{6}{c}{Adam} \\
     Learning rate of actors \& critics & \multicolumn{6}{c}{$3e^{-4}$} \\
     Discount rate ($\gamma$) & \multicolumn{6}{c}{0.99} \\
     Target smoothing coefficient ($\rho$) & \multicolumn{6}{c}{0.005} \\
     Replay ratio (gradient step) & \multicolumn{6}{c}{1} \\
     Mini-batch size & \multicolumn{6}{c}{256} \\
     \midrule
     Diversity temperature ($\beta$) & \multicolumn{2}{c|}{$0.8$} & \multicolumn{4}{c}{N/A} \\
     \cline{2-7}
     Target entropy ($\mathcal{\bar H}$) & \multicolumn{2}{c|}{N/A} & $0.5|\A|$ & \multicolumn{2}{c|}{N/A} & $0.5|\A|$ \\
     Learning rate for $\alpha$ auto-adjustment & \multicolumn{2}{c|}{$5e^{-3}$} & $3e^{-2}$ & \multicolumn{2}{c|}{N/A} & $3e^{-4}$ \\
     \cline{2-7}
     Entropy coefficient ($\alpha$) & \multicolumn{3}{c|}{N/A} & 0.3 & 0.7 & N/A \\
     \midrule
     Number of diversity estimation pairs ($n$) & \multicolumn{2}{c|}{$8$} & \multicolumn{4}{c}{N/A} \\
     \cline{2-7}
     Number of SVGD particles ($k$) & \multicolumn{3}{c|}{N/A} & $10$ & $32$ & N/A \\
     \midrule
     Policy delay (actor update period) & \multicolumn{2}{c|}{N/A, equivalent to 1} & 2 & \multicolumn{3}{c}{N/A, equivalent to 1} \\
    \bottomrule
\end{tabular}

    }
    \label{tab:sharedparam}
\end{table}

\begin{table}[t]
    \centering
    \caption{Temperature settings for MuJoCo}
    \small
    \begin{tabular}{c|c|c|c|c|c}
    \toprule
     \textbf{DrAmort} & \textbf{DrDiffus} & DACER & SQL & S$^2$AC & SAC \\
    \midrule
    $0.5$ & $0.2$ & $-0.9|\A|$ \cite{wang2024diffusion} & \multicolumn{2}{c|}{$0.2$ for Ant, $1.0$ for the rest \cite{messaoud2024s}} & $-|\A|$ \\
    \bottomrule
\end{tabular}

    \label{tab:mujoco-temps}
\end{table}

Regarding compute resources, all of our experiments are conducted with a Linux server with 8 NVIDIA RTX 3090 GPUs and an Intel Xeon Platinum 8375C CPU. All trainings are done with one GPU.

\subsection{Relation to Amortized Inference and Traditional Gaussian Actor}
The formulation of stochastic mapping actors is closed to the definition of amortized inference. Due to that the term ``amortized actor'' has been used in \cite{haarnoja2017reinforcement} and adopted in \cite{messaoud2024s}, we name our formulation as ``stochastic mapping actors''.

The formulation of stochastic mapping actors is also compatible with traditional Gaussian actor, by treating $f(s, z) \equiv \mu_\theta (s) + \sigma_\theta z$ and $z \sim \gaus (\boldsymbol{0}, I)$. Our method is also directly applicable to Gaussian actor.

\newpage
\section{Additional Results}
\subsection{Computational costs}
We record the average time cost per 1000 training steps in Ant-v4 environment. The test is conducted on a Linux server with 8 NVIDIA RTX 3090 GPUs and an Intel Xeon Platinum 8375C CPU. For each algorithm, we run three trials on GPU cards 0, 1 and 2, respectively, and then average the training time per 1000 steps. All algorithms are implemented with PyTorch 2.5.1, and the CUDA version is 12.5.
\begin{table}[h]
    \centering
    \caption{Computational costs of all compared algorithms}
    \begin{tabular}{c|cccccc}
        \toprule
        & SAC & SQL & \textbf{DrAmort} & DACER & \textbf{DrDiffus} & S$^2$AC \\
        \midrule
         Training time per 1000 steps & 22.7s & 18.9s & 21.0s & 76.9s & 84.8s & 220.5s \\
         Peak GPU memory usage & 352M & 472M & 392M & 364M & 608M & 3462M \\
        \bottomrule
    \end{tabular}
    \label{tab:placeholder}
\end{table}

According to the table, we found that DrAmort is even faster than SAC. The reason is that SAC computes and back-propagates the log-probability at certain actions, which requires more sequential tensor operations. Though DrAmort requires samples, the computation is parallelized and takes fewer sequential tensor operations. As a result, DrAmort uses fewer GPU clock ticks and is slightly faster than SAC, given that the neural network is sufficiently small compared to the GPU’s parallel computation capability. For DrDiffus, it takes about 10\% more time for training compared to DACER, but its diversity in multi-goal PointMaze is better.

For large-scale tasks that require larger networks, such as visual control, we can inject the latent variable at the last few layers. For example, if a CNN is used, the latent variables can be injected after the convolutional layers. In this case, the computation time and memory usage will not significantly increase.

\subsection{Trajectories in Medium and Hard Multi-goal PointMaze \label{sec:moretrajs}}

\begin{figure}[h]
    \centering
    \includegraphics[width=\linewidth]{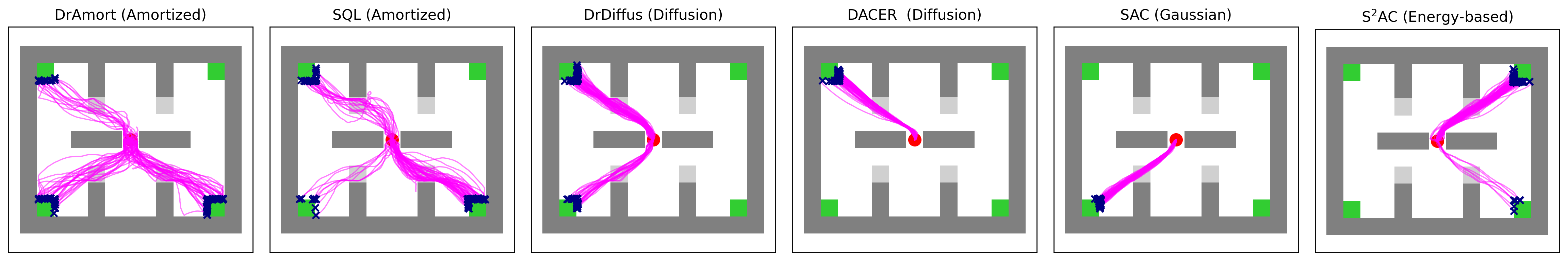}
    \caption{Evaluation trajectories of all tested algorithms in the medium maze.}
    \label{fig:medium-trajs}
\end{figure}

\begin{figure}[h]
    \centering
    \includegraphics[width=\linewidth]{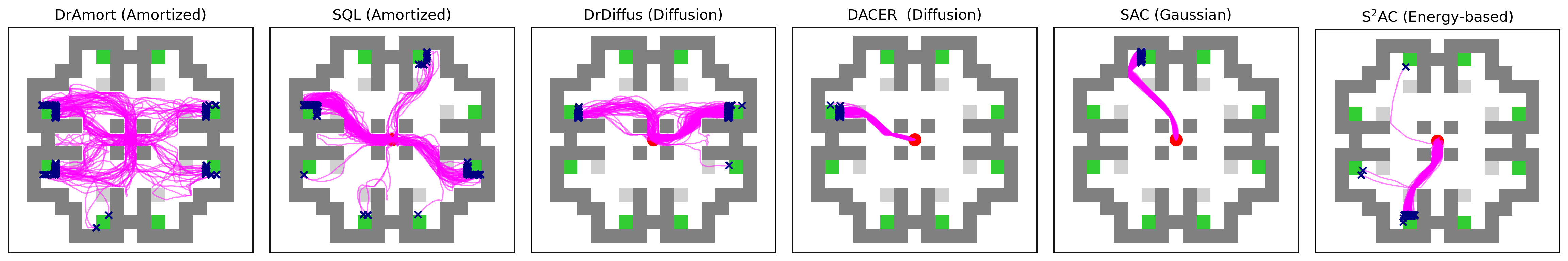}
    \caption{Evaluation trajectories of all tested algorithms in the hard maze.}
    \label{fig:hard-trajs}
\end{figure}

Figs. \ref{fig:medium-trajs} and \ref{fig:hard-trajs} demonstrate the trajectories in medium and hard mazes, respectively.

Consistently, DrAmort exhibits the best diversity.

\newpage

\subsection{Sensitivity of Temperature Hyperparameter in Multi-goal PointMaze}
We display the learning curves of all evaluated algorithms with varied temperatures. 
For our DrAmort and DrDiffus, we show a sensitivity study with five different temperatures.
For baseline algorithms, we present three temperatures to explain our choice of their temperatures in our comparison study. Specifically, for all baseline algorithms, using a higher temperature than our choice leads to suboptimal success rate in at least one maze; while using a lower one leads to lower diversity.

\begin{figure}[H]
    \centering
    \includegraphics[width=\linewidth]{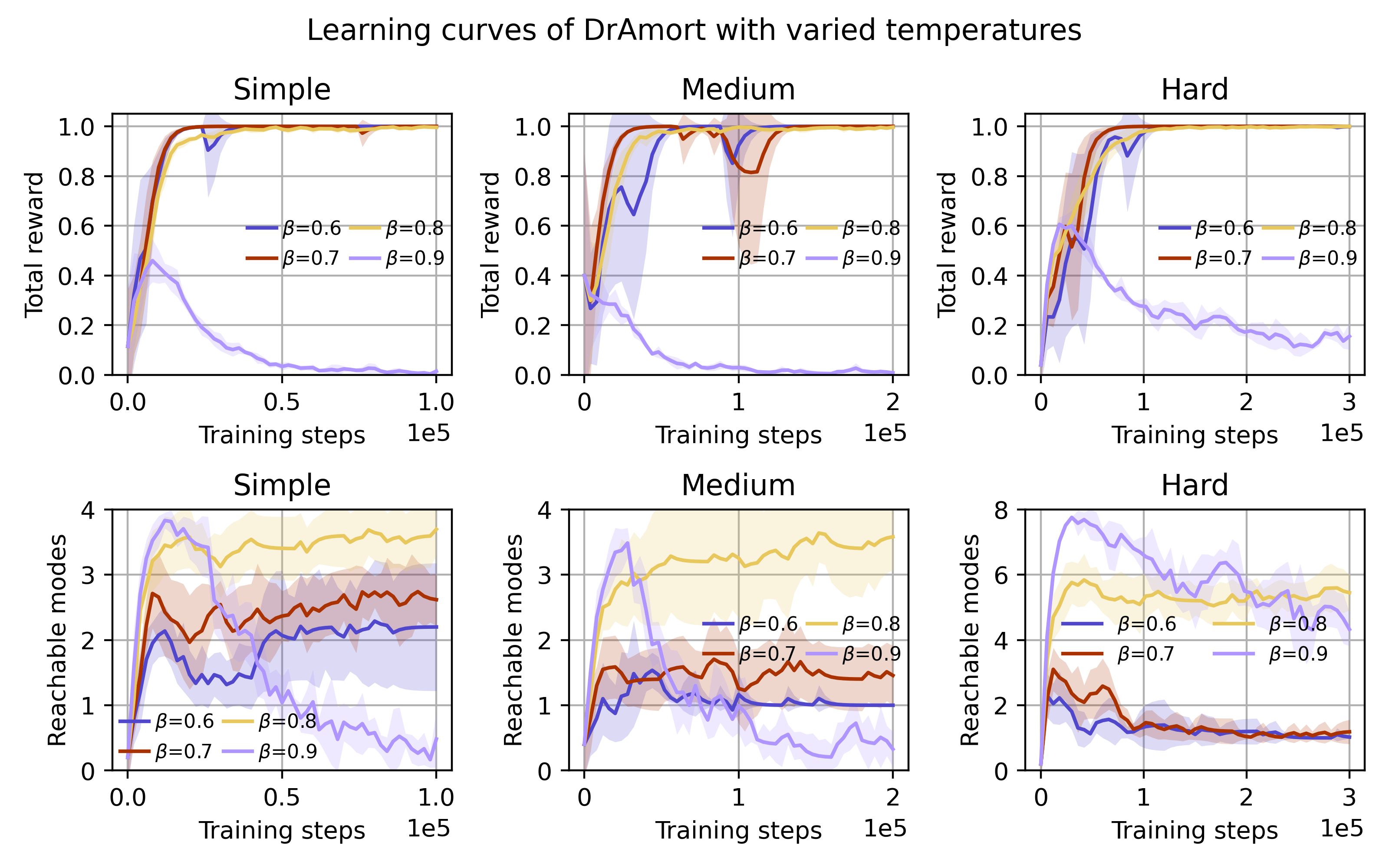}
    \caption{Learning curves of DrAmort in multi-goal PointMaze with varied temperatures.}
\end{figure}

\begin{figure}[H]
    \centering
    \includegraphics[width=\linewidth]{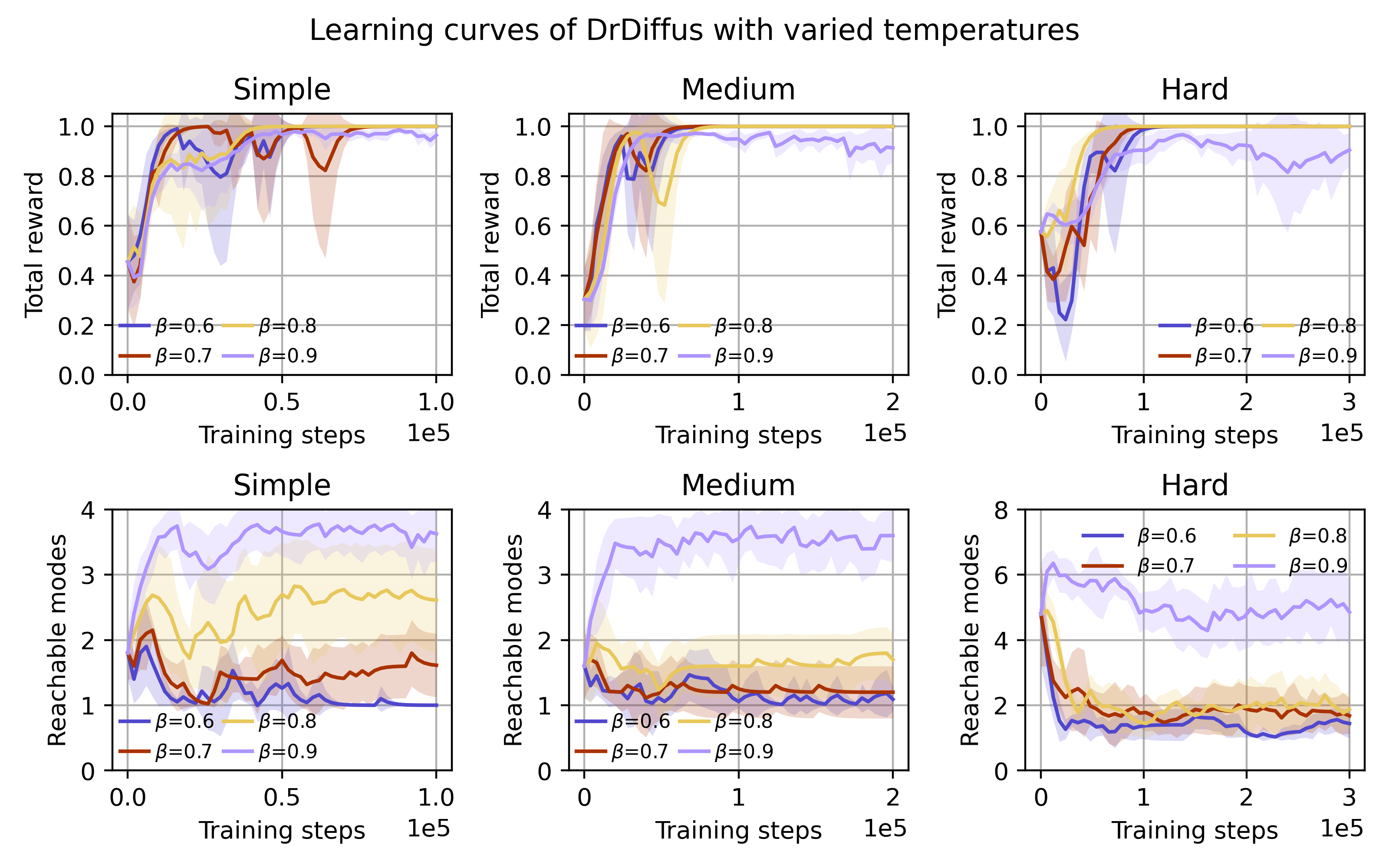}
    \caption{Learning curves of DrDiffus in multi-goal PointMaze with varied temperatures.}
\end{figure}

\begin{figure}[H]
    \centering
    \includegraphics[width=\linewidth]{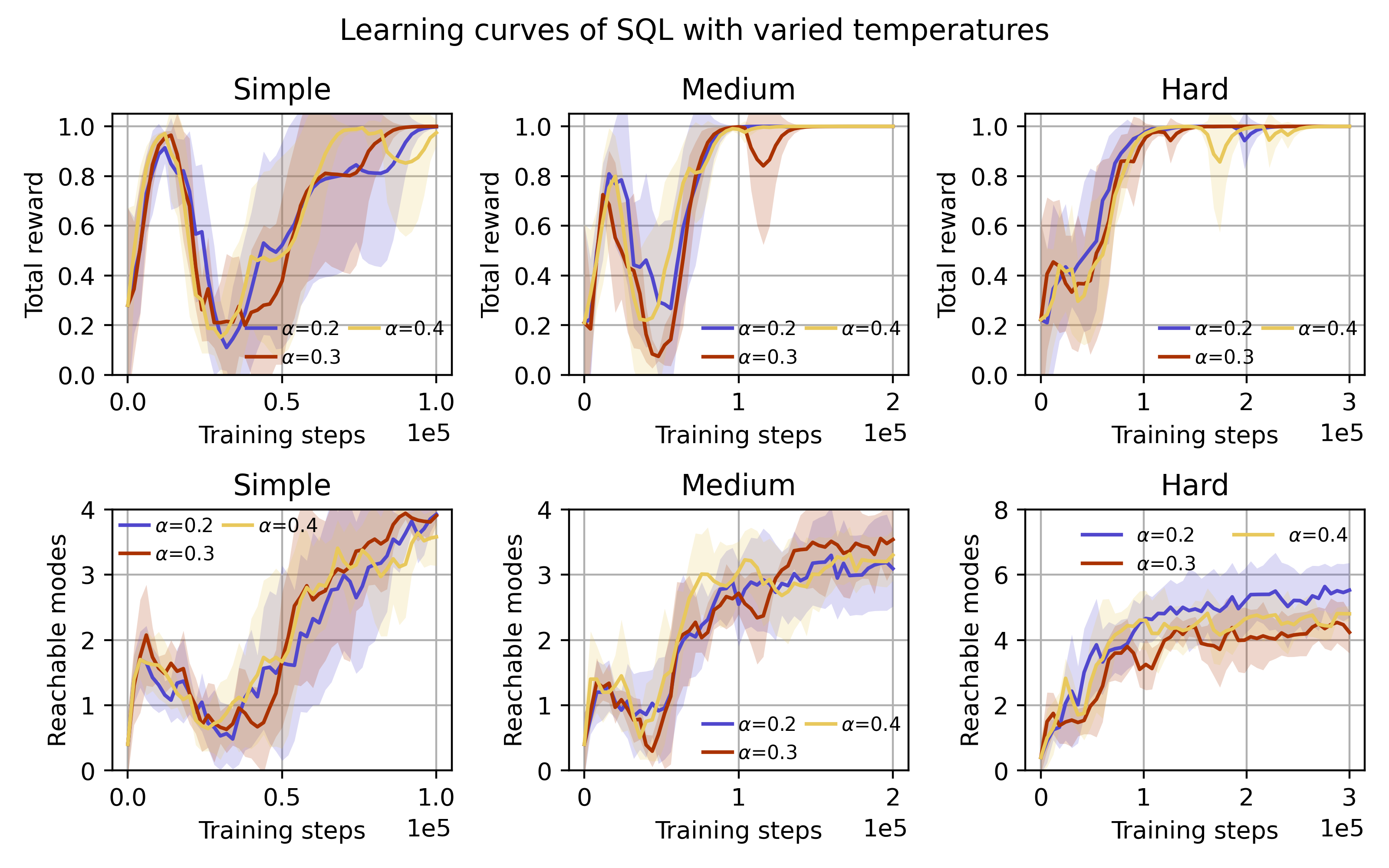}
    \caption{Learning curves of SQL in multi-goal PointMaze with varied temperatures.}
\end{figure}

\begin{figure}[H]
    \centering
    \includegraphics[width=\linewidth]{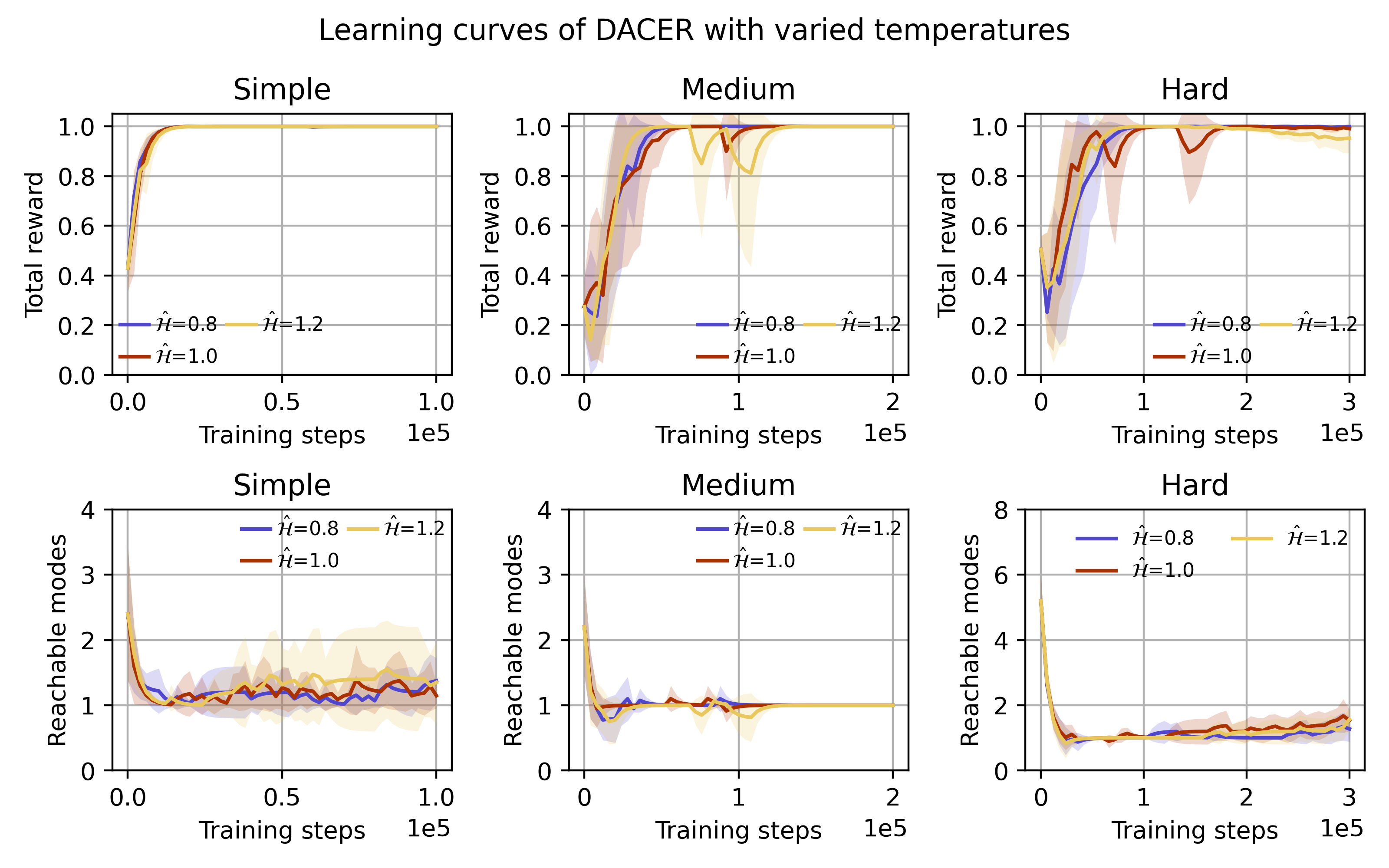}
    \caption{Learning curves of DACER in multi-goal PointMaze with varied temperatures.}
\end{figure}

\begin{figure}[H]
    \centering
    \includegraphics[width=\linewidth]{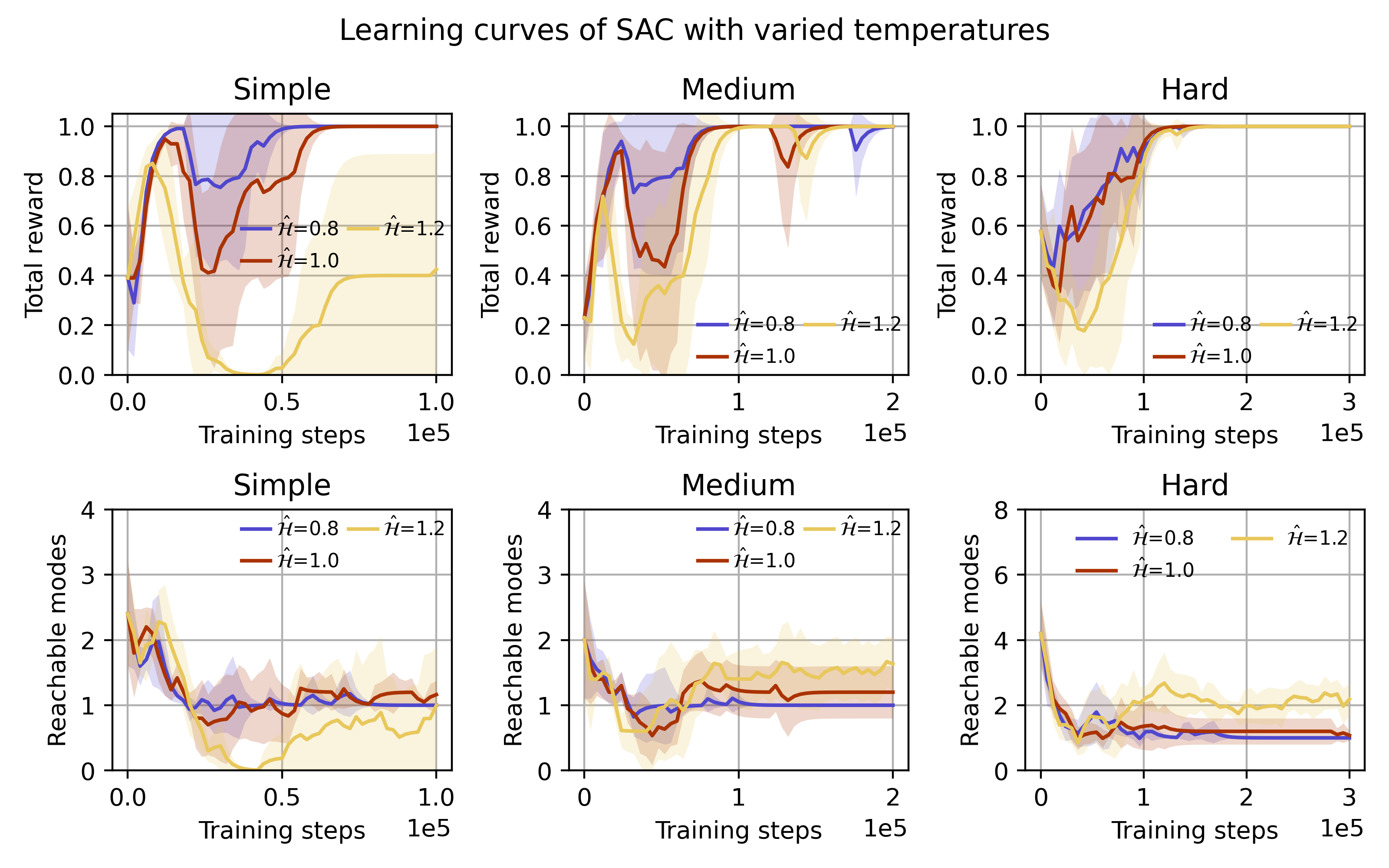}
    \caption{Learning curves of SAC in multi-goal PointMaze with varied temperatures.}
\end{figure}
\begin{figure}[H]
    \centering
    \includegraphics[width=\linewidth]{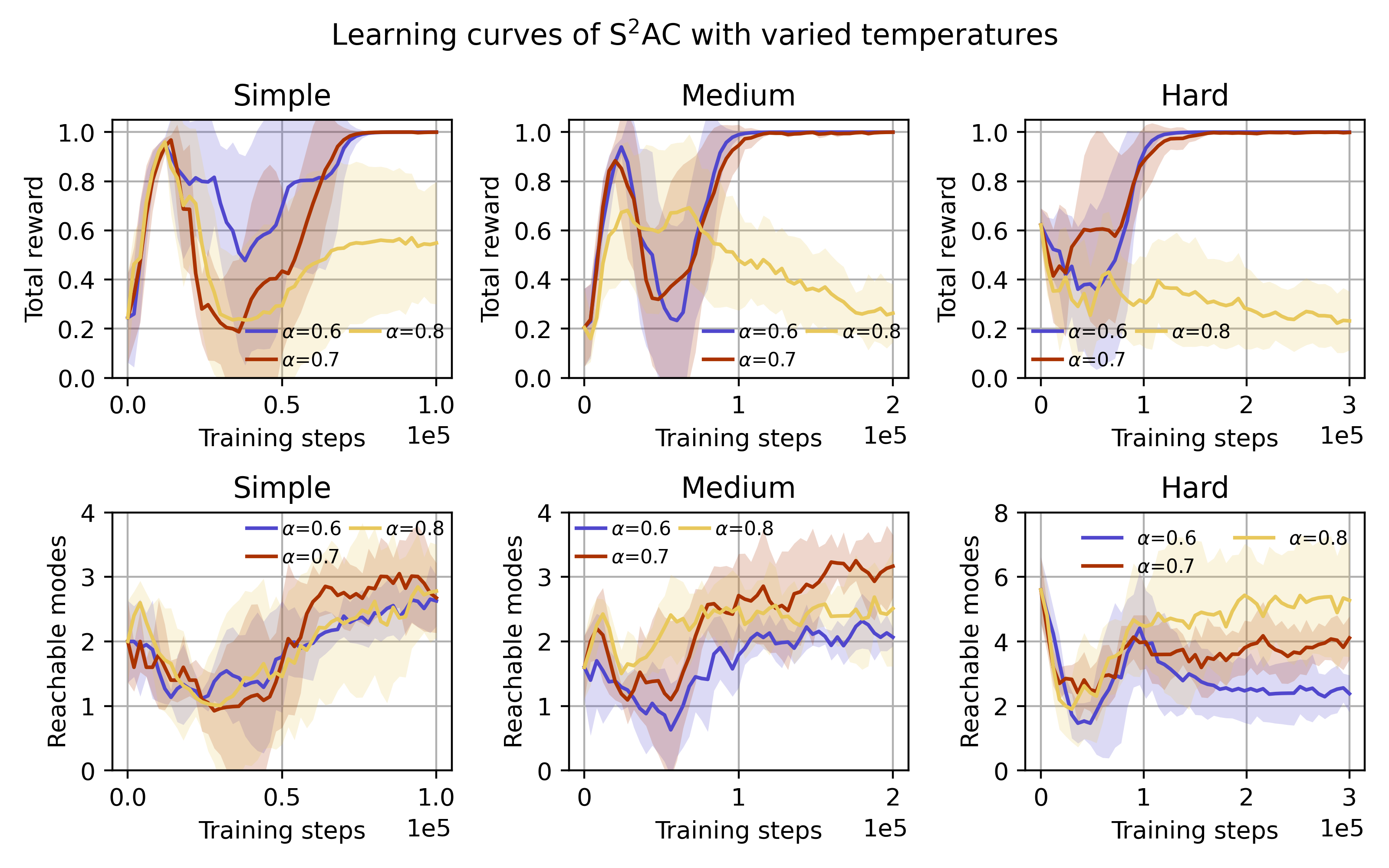}
    \caption{Learning curves of S$^2$AC in multi-goal PointMaze with varied temperatures.}
\end{figure}

\newpage

\subsection{Sensitivity of Temperature in Game Content Generation}
\begin{figure}[!h]
    \centering
    \includegraphics[width=\linewidth]{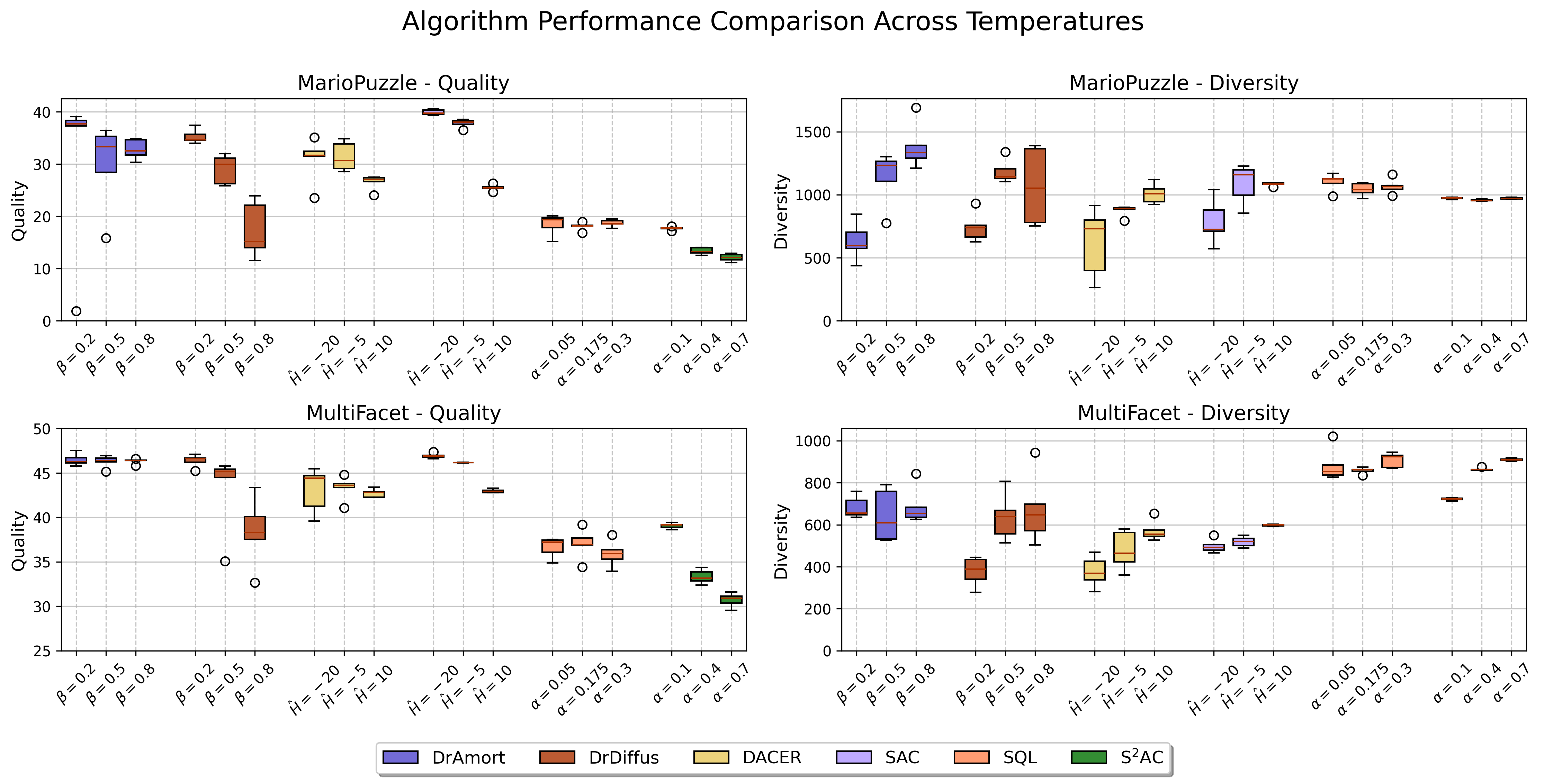}
        \caption{Sensitivity of temperature across tasks and metrics. The ticks of $x$-axis details the temperature.}
    \label{fig:smb-abl}
\end{figure}

We train each algorithm with three groups of temperatures, namely high, mid, low, to investigate the sensitivity of temperature. 
The high temperature group uses the same temperature with the multi-goal PointMaze. The low temperature group are set to $-|\A|$ for SAC and DACER and near zero for DrAC, SQL and S$^2$AC. The mid temperature group use temperatures averaged between low and high groups.
Fig. \ref{fig:smb-abl} shows the algorithm performance across temperatures.

Generally, the quality decreases along with raising temperature while diversity increases along with raising temperature. To provide an overall evaluation, we treat all policies trained by each algorithm across temperatures and seeds as a population, and compute the hypervolume (HV) metric for these populations \cite{zitzler2002multiobjective}. The reference point is set according to the lowest quality and diversity performance across all algorithms, temperatures and seeds.
\begin{figure}[!h]
    \centering
    \includegraphics[width=\linewidth]{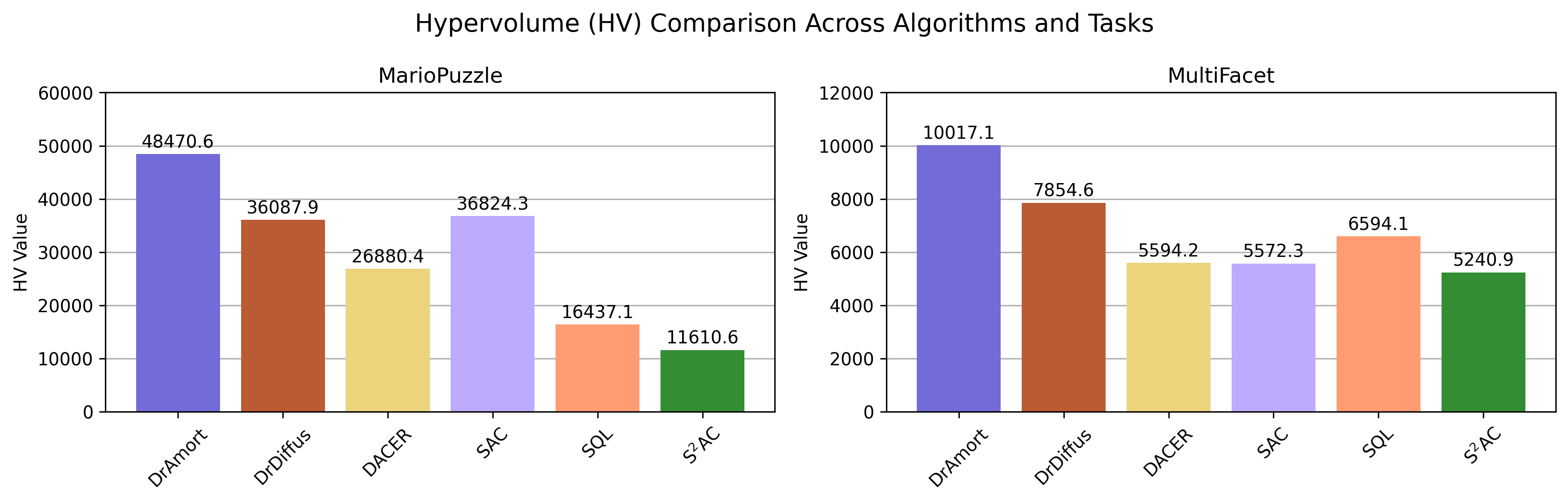}
    \caption{HV comparison of policy populations obtained by training each algorithm through three different temperatures.}
    \label{fig:smb-hv}
\end{figure}

The results show DrAmort produces the best population in terms of the HV metric, DrDiffus outperform baselines in MultiFacet and shows competitive performance in MarioPuzzle.
\newpage

\subsection{t-SNE visualization of Generated Game Levels\label{sec:tsne}}
\begin{figure}[H]
    \centering
    \includegraphics[width=0.9\linewidth]{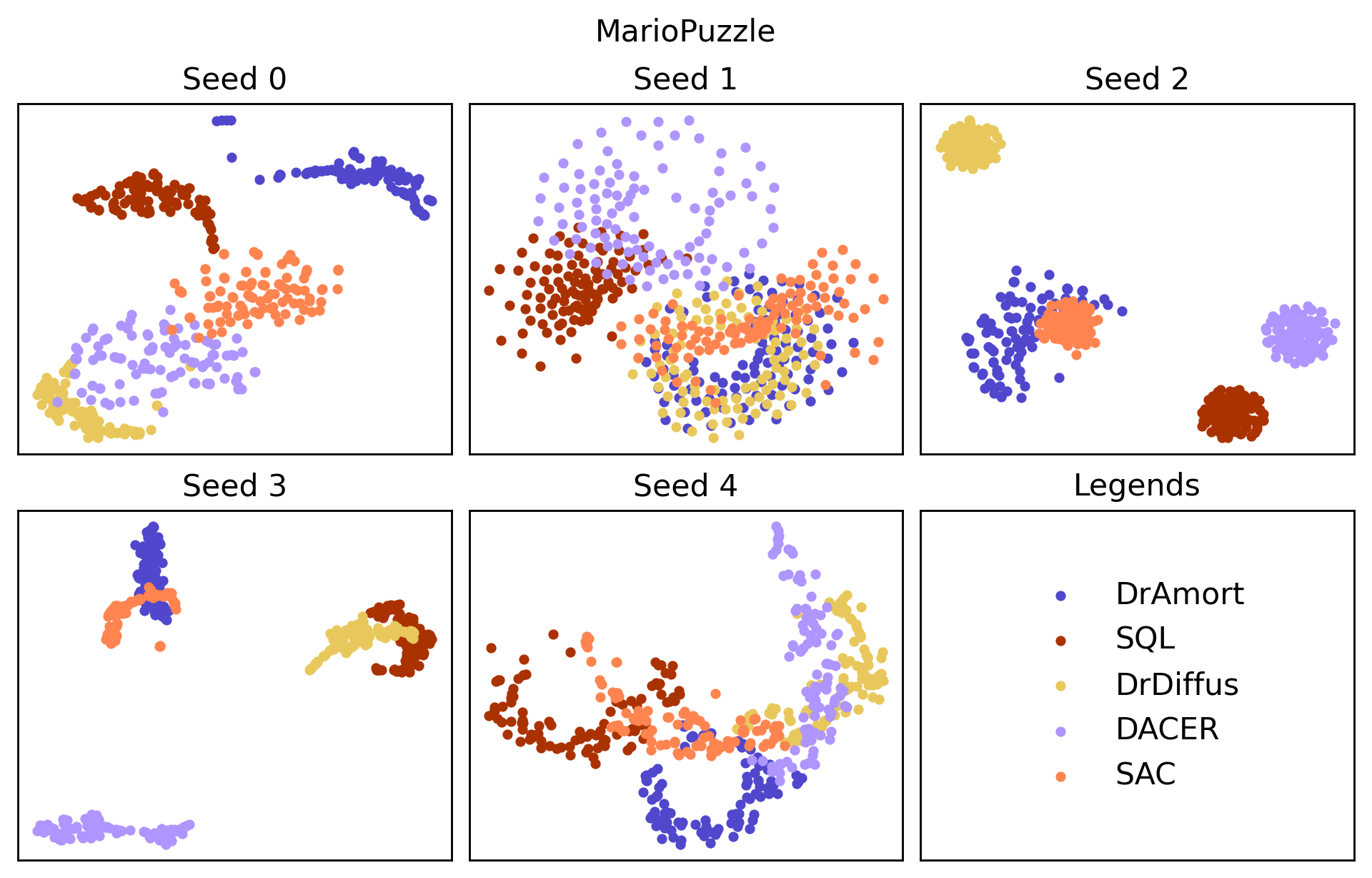}
    \caption{t-SNE embeddings of levels generated by each algorithm, under MarioPuzzle style.}
    \label{fig:tsnemp}
\end{figure}

\begin{figure}[H]
    \centering
    \includegraphics[width=0.9\linewidth]{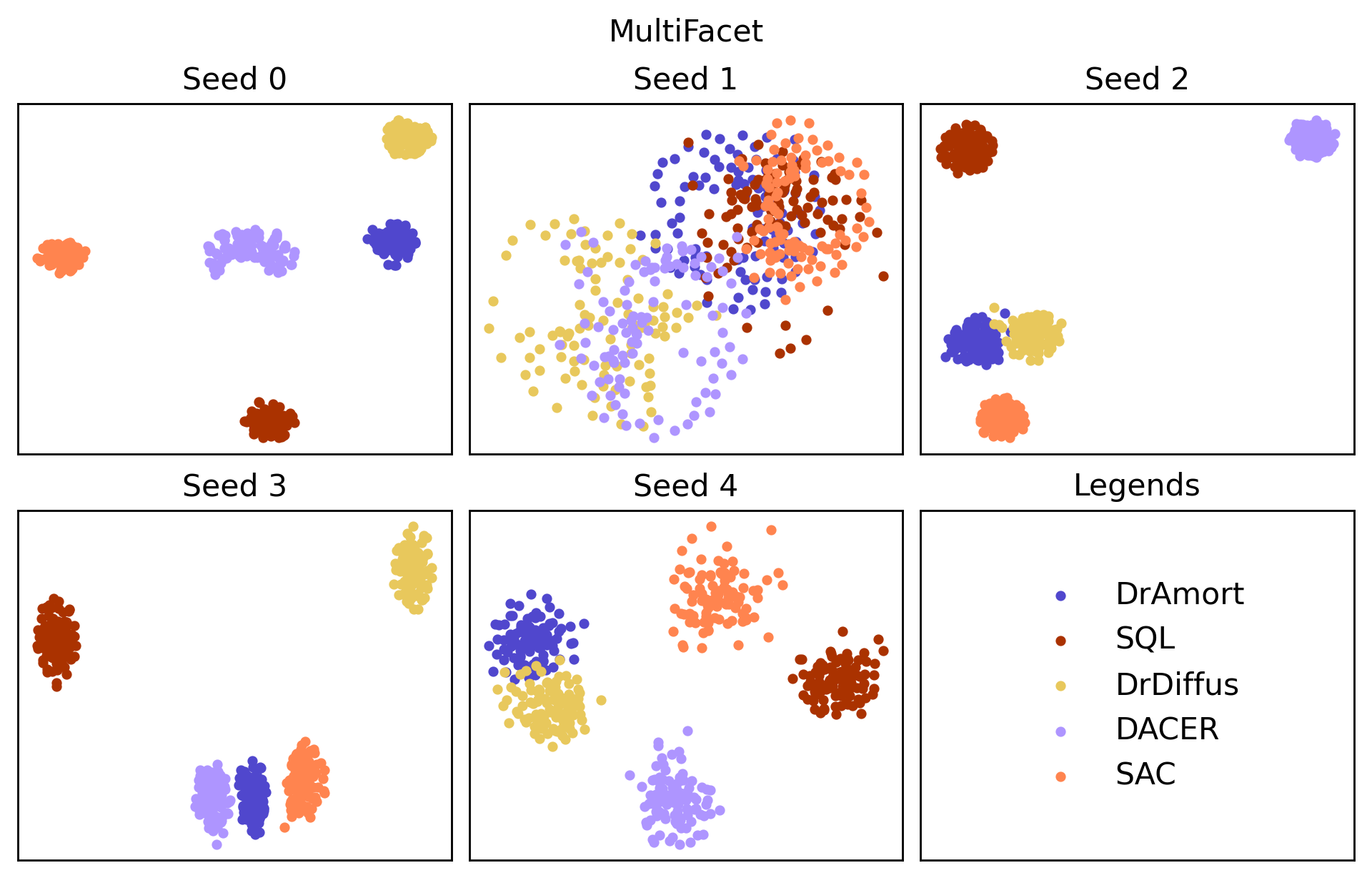}
    \caption{t-SNE embeddings of levels generated by each algorithm, under MultiFacet style.}
    \label{fig:tsnemf}
\end{figure}

To compare the diversity of levels generated by each algorithm more intuitively, we embed levels generated by each algorithm with t-SNE \cite{van2008visualizing}, and plot the embeddings in Fig. \ref{fig:tsnemp} and Fig. \ref{fig:tsnemf}.

Observing the embeddings, differences in terms of diversity are not significant. Different algorithms generally form different clusters. This may indicate that there is still ample room to improve the diversity with little harm to quality. Future work may investigate how to unleash the expressivity of multimodal policies more completely. Besides, average pairwise Hamming distance is a vanilla method to measure diversity of game content, more in-depth quantitative analysis may be considered in future work.

\newpage
\subsection{Sensitivity of Temperature Hyperparameter in MuJoCo}
We demonstrate the sensitivity of temperature hyperparameter of DrAmort and DrDiffus as follows.

\begin{figure}[!h]
    \centering
    \includegraphics[width=\linewidth]{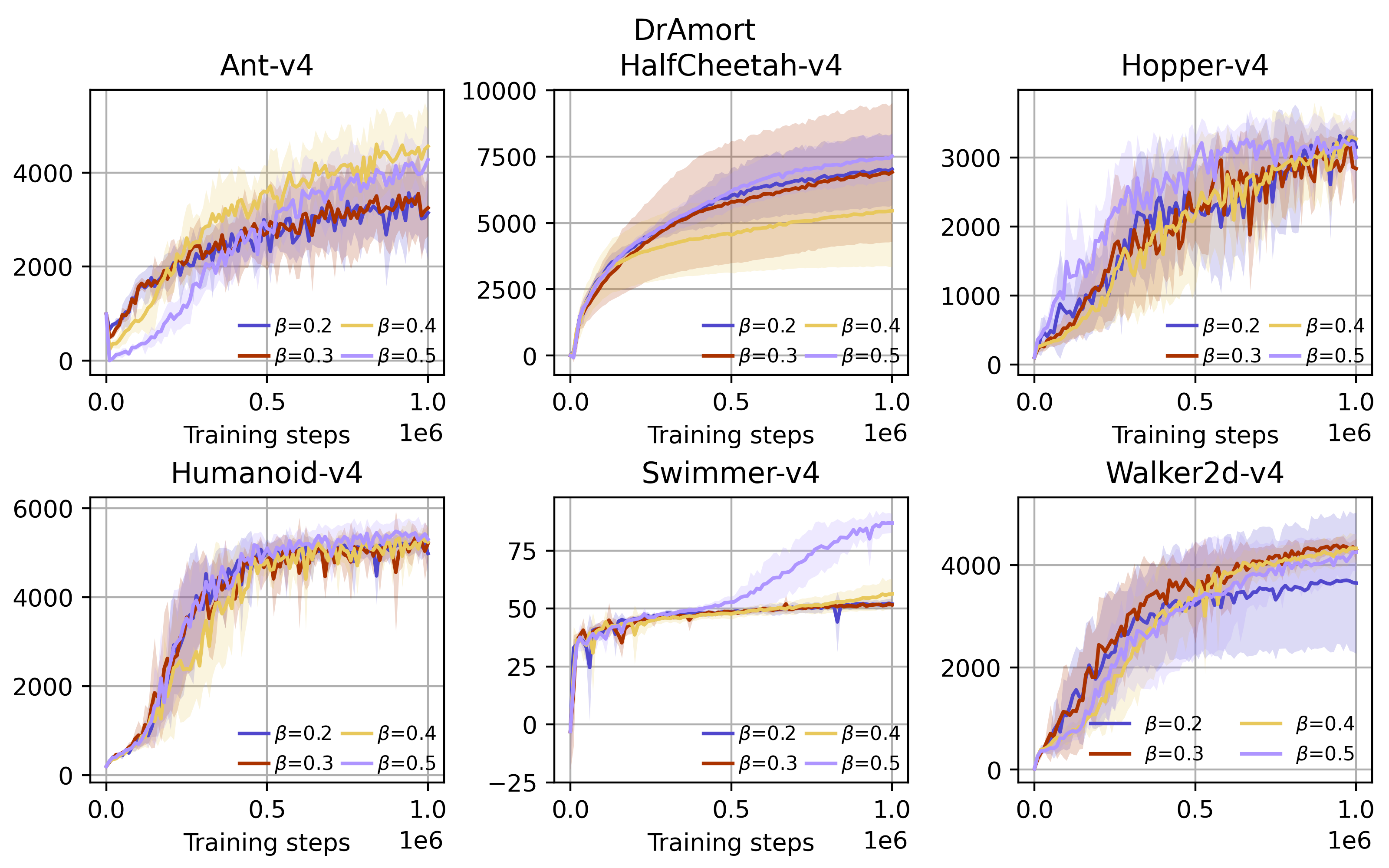}
    \caption{Learning curves of DrAmort with varied temperatures in MuJoCo}
    \label{fig:mujoco-dramt}
\end{figure}
\vspace{-6pt}
\begin{figure}[!h]
    \centering
    \includegraphics[width=\linewidth]{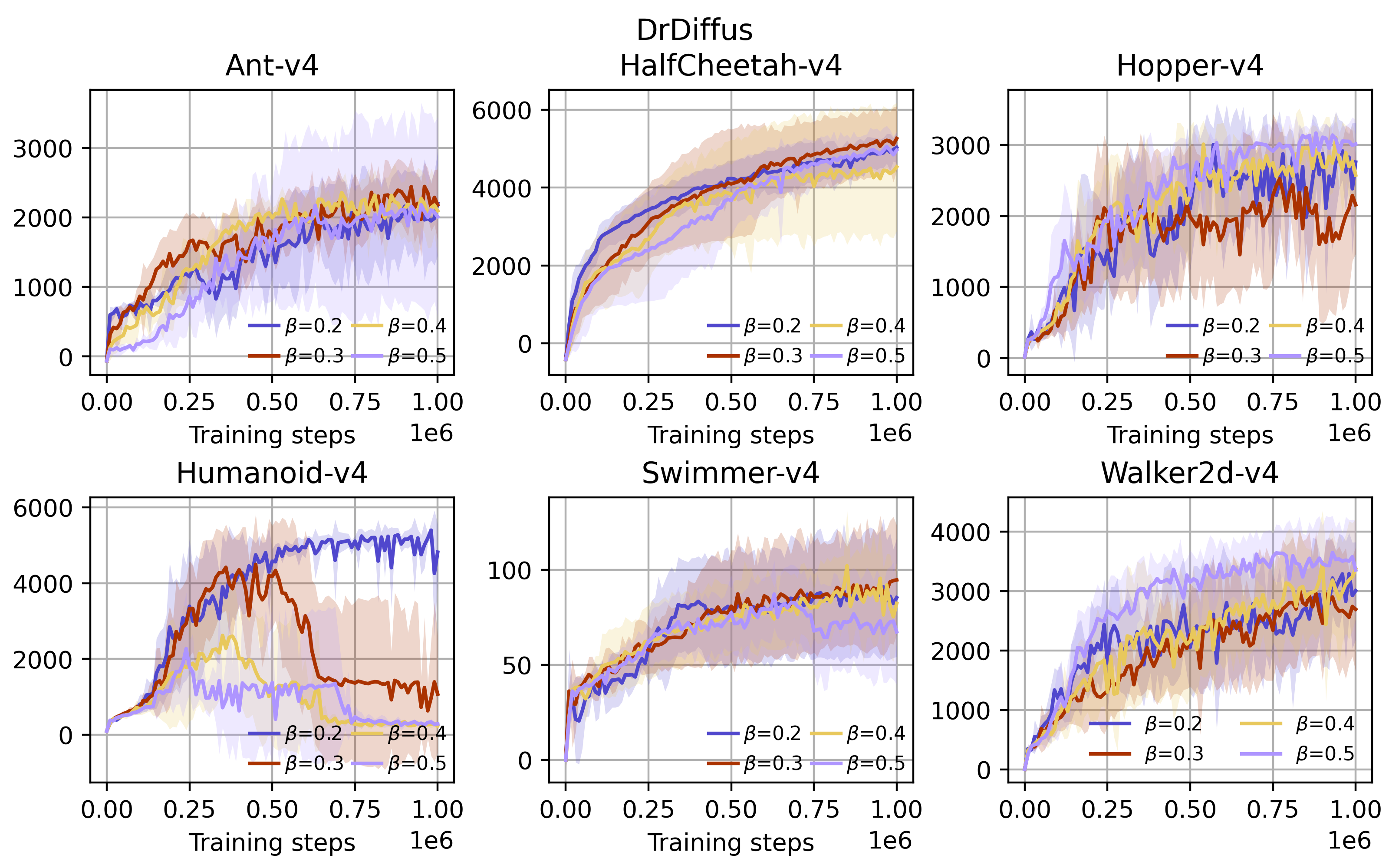}
    \caption{Learning curves of DrDiffus with varied temperatures in MuJoCo}
    \label{fig:mujoco-drdfs}
\end{figure}

DrAmort is robust to the temperature, overall $\beta=0.5$ deliveries superior performance. DrDiffus suffers from a performance drop in Humanoid with $\beta > 0.2$. More investigation into this phenomenon is an important future work.

\newpage
\section*{NeurIPS Paper Checklist}
\begin{enumerate}

\item {\bf Claims}
    \item[] Question: Do the main claims made in the abstract and introduction accurately reflect the paper's contributions and scope?
    \item[] Answer: \answerYes 
    \item[] Justification: The abstract and introduction accurately describe the proposed algorithm (policy gradient via reparameterization + distance-based diversity regularization) and experimental scope (multi-goal achieving, generative RL, MuJoCo). The abstract and the introduction align with the technical content in the paper (Section \ref{sec:method}, \ref{sec:experiments}, Appendix \ref{sec:code}).

\item {\bf Limitations}
    \item[] Question: Does the paper discuss the limitations of the work performed by the authors?
    \item[] Answer: \answerYes
    \item[] Justification: The conclusion (Section \ref{sec:conclusion}) involves a discussion of limitations. 

\item {\bf Theory assumptions and proofs}
    \item[] Question: For each theoretical result, does the paper provide the full set of assumptions and a complete (and correct) proof?
    \item[] Answer: \answerNA{} 
    \item[] Justification: 

\item {\bf Experimental result reproducibility}
    \item[] Question: Does the paper fully disclose all the information needed to reproduce the main experimental results of the paper to the extent that it affects the main claims and/or conclusions of the paper (regardless of whether the code and data are provided or not)?
    \item[] Answer: \answerYes 
    \item[] Justification: Benchmarks and tested algorithms are described in Section \ref{sec:experiments}. Experiment settings, including evaluation criteria, are described within Section \ref{sec:experiments}. Hyperparameters are detailed in Appendix \ref{sec:details}. Appendix \ref{sec:code} provides detailed pseudo-code of the proposed algorithm. Our code will be released after acceptance.

\item {\bf Open access to data and code}
    \item[] Question: Does the paper provide open access to the data and code, with sufficient instructions to faithfully reproduce the main experimental results, as described in supplemental material?
    \item[] Answer: \answerYes 
    \item[] Justification: We will submit our code in supplemental materials, and we will release our code with instructions after acceptance. Our code contains training and evaluation. Customized RL environments are also included with the necessary assets. We do not use training data.

\item {\bf Experimental setting/details}
    \item[] Question: Does the paper specify all the training and test details (e.g., data splits, hyperparameters, how they were chosen, type of optimizer, etc.) necessary to understand the results?
    \item[] Answer: \answerYes 
    \item[] Justification: Experiment settings and evaluation criteria are introduced within the Section \ref{sec:experiments}. Hyperparameters, including network architecture and the type of optimizer, are detailed in Appendix \ref{sec:hyperparams}.

\item {\bf Experiment statistical significance}
    \item[] Question: Does the paper report error bars suitably and correctly defined or other appropriate information about the statistical significance of the experiments?
    \item[] Answer: \answerYes 
    \item[] Justification: We repeat all training with five seeds, and all results are reported with mean and standard deviation across the five seeds. All algorithms used the same set of seeds. All evaluations for all checkpoints are repeated at least $20$ times to average. 

\item {\bf Experiments compute resources}
    \item[] Question: For each experiment, does the paper provide sufficient information on the computer resources (type of compute workers, memory, time of execution) needed to reproduce the experiments?
    \item[] Answer: \answerYes 
    \item[] Justification: We describes our compute resources in Appendix \ref{sec:hyperparams}.
    
\item {\bf Code of ethics}
    \item[] Question: Does the research conducted in the paper conform, in every respect, with the NeurIPS Code of Ethics \url{https://neurips.cc/public/EthicsGuidelines}?
    \item[] Answer: \answerYes 
    \item[] Justification: The research involves algorithmic development and evaluation on simulation benchmarks and open-source game benchmarks. It does not involve human subjects or obviously ethically sensitive applications, and we expect that it conforms to the NeurIPS Code of Ethics.

\item {\bf Broader impacts}
    \item[] Question: Does the paper discuss both potential positive societal impacts and negative societal impacts of the work performed?
    \item[] Answer: \answerNA 
    \item[] Justification: This research focuses on basic deep RL algorithm development, and all experiments are conducted in simulated imaginary environments or digital games without any sensitive subjects. We expect there is no societal impact.

\item {\bf Safeguards}
    \item[] Question: Does the paper describe safeguards that have been put in place for responsible release of data or models that have a high risk for misuse (e.g., pretrained language models, image generators, or scraped datasets)?
    \item[] Answer: \answerNA 
    \item[] Justification: This work focuses on algorithm development and experiments are conducted with simulated environments and open-source games, so we expect the paper poses no such risks.

\item {\bf Licenses for existing assets}
    \item[] Question: Are the creators or original owners of assets (e.g., code, data, models), used in the paper, properly credited and are the license and terms of use explicitly mentioned and properly respected?
    \item[] Answer: \answerYes
    \item[] Justification: Our experiments dependent on open-source assets in D4RL \cite{fu2020d4rl} and a game content generation \cite{wang2024negatively}, which are credited by citations in Section \ref{sec:experiments}.

\item {\bf New assets}
    \item[] Question: Are new assets introduced in the paper well documented and is the documentation provided alongside the assets?
    \item[] Answer: \answerNA 
    \item[] Justification: This work does not involve new assets.

\item {\bf Crowdsourcing and research with human subjects}
    \item[] Question: For crowdsourcing experiments and research with human subjects, does the paper include the full text of instructions given to participants and screenshots, if applicable, as well as details about compensation (if any)? 
    \item[] Answer: \answerNA 
    \item[] Justification: This work does not involve any crowdsourcing experiment nor research with human subjects. 

\item {\bf Institutional review board (IRB) approvals or equivalent for research with human subjects}
    \item[] Question: Does the paper describe potential risks incurred by study participants, whether such risks were disclosed to the subjects, and whether Institutional Review Board (IRB) approvals (or an equivalent approval/review based on the requirements of your country or institution) were obtained?
    \item[] Answer: \answerNA 
    \item[] Justification: This work does not involve any crowdsourcing experiment nor research with human subjects.

    \item {\bf Declaration of LLM usage}
    \item[] Question: Does the paper describe the usage of LLMs if it is an important, original, or non-standard component of the core methods in this research? Note that if the LLM is used only for writing, editing, or formatting purposes and does not impact the core methodology, scientific rigorousness, or originality of the research, declaration is not required.
    \item[] Answer: \answerNA 
    \item[] Justification: 
    This work does not involve LLMs as any important, original, or non-standard components.
\end{enumerate}

\end{document}